\RequirePackage{float}
\documentclass[pdflatex,sn-basic,Numbered]{sn-jnl}
\usepackage{graphicx}
\usepackage{tikz}
\usetikzlibrary{arrows.meta,positioning}
\usepackage{threeparttable}
\usepackage{multirow}
\usepackage{amsmath,amssymb,amsfonts}
\usepackage[title]{appendix}
\usepackage{xcolor}
\usepackage{colortbl}
\definecolor{methodshade}{gray}{0.94}
\usepackage{textcomp}
\usepackage{booktabs}
\usepackage{array}
\usepackage{microtype}
\usepackage{placeins}
\usepackage{algorithm}
\usepackage{algorithmicx}
\usepackage{algpseudocode}

\hypersetup{
  pdftitle={Inverse-LLaVA: Rethinking Multimodal Alignment via Text-to-Vision Mapping},
  pdfauthor={Xuhui Zhan and Tyler Derr},
  pdfsubject={Text-to-vision fusion for vision-language models},
  pdfkeywords={Inverse-LLaVA, multimodal learning, text-to-vision mapping}
}

\begin{document}

\title[Inverse-LLaVA]{Inverse-LLaVA: Rethinking Multimodal Alignment via Text-to-Vision Mapping}

\author[1]{\fnm{Xuhui} \sur{Zhan}}\email{xuhui.zhan@vanderbilt.edu}

\author*[1,2]{\fnm{Tyler} \sur{Derr}}\email{tyler.derr@vanderbilt.edu}

\affil*[1]{\orgdiv{Data Science Institute}, \orgname{Vanderbilt University}, \orgaddress{\street{2201 West End Ave}, \city{Nashville}, \postcode{37235}, \state{Tennessee}, \country{United States}}}

\affil[2]{\orgdiv{Computer Science Department}, \orgname{Vanderbilt University}, \orgaddress{\street{2201 West End Ave}, \city{Nashville}, \postcode{37235}, \state{Tennessee}, \country{United States}}}

\abstract{
Connecting pretrained vision and language models usually involves projecting image features into the language model's input space. Inverse-LLaVA reverses this mapping within decoder attention: language states are projected to the visual feature dimension, and modality-specific maps produce residual query, key, and value updates. Fusion and low-rank adaptation (LoRA) learn jointly from 665K visual instructions, with frozen backbones and no separate alignment stage. Across nine primary benchmark evaluations, the final 7B model approaches two-stage LLaVA-1.5 on several tasks. It scores 78.45\% on VQAv2 versus 79.13\% for official LLaVA-LoRA, and 50.96\% versus 48.56\% on VizWiz; TextVQA is lower at 56.96\% versus 58.47\%. Controlled studies examine fusion components, insertion depth, visual features, and language-model size. Representation analysis shows that the text maps preserve much of the pairwise similarity ordering while changing its geometric spread. Additional paired supervision improves celebrity recognition, while instruction replay repairs caption-induced answer-format failures. Analytical cost expressions and fixed-work profiles separate the additional fusion computation from the omitted alignment stage. These findings establish text-to-vision attention fusion as a practical alternative for instruction-only multimodal adaptation.}

\keywords{Single-stage Multimodal Learning; Text-to-Vision Mapping; Cross-modal Fusion; Vision--Language Models}

\maketitle

\section{Introduction}
\label{sec:intro}

Vision--language models combine pretrained visual representations with the generative capabilities of language models. Flamingo~\citep{alayrac2022flamingovisuallanguagemodel}, BLIP-2~\citep{li2023blip2bootstrappinglanguageimagepretraining}, and LLaVA~\citep{liu2023llava,liu2023improvedllava} connect these components in different ways. LLaVA-1.5 uses a multilayer perceptron (MLP) to project visual features to the language model's hidden space.

The usual visual-projector design couples two choices: visual features are mapped to the language width, and modalities share the decoder input interface. We ask whether the connection can instead be constructed inside attention, with language states first mapped to the visual feature width. This changes the modality-specific transformations while preserving the decoder's pretrained residual paths.

\begin{figure*}[!ht]
\centering
\includegraphics[width=\textwidth]{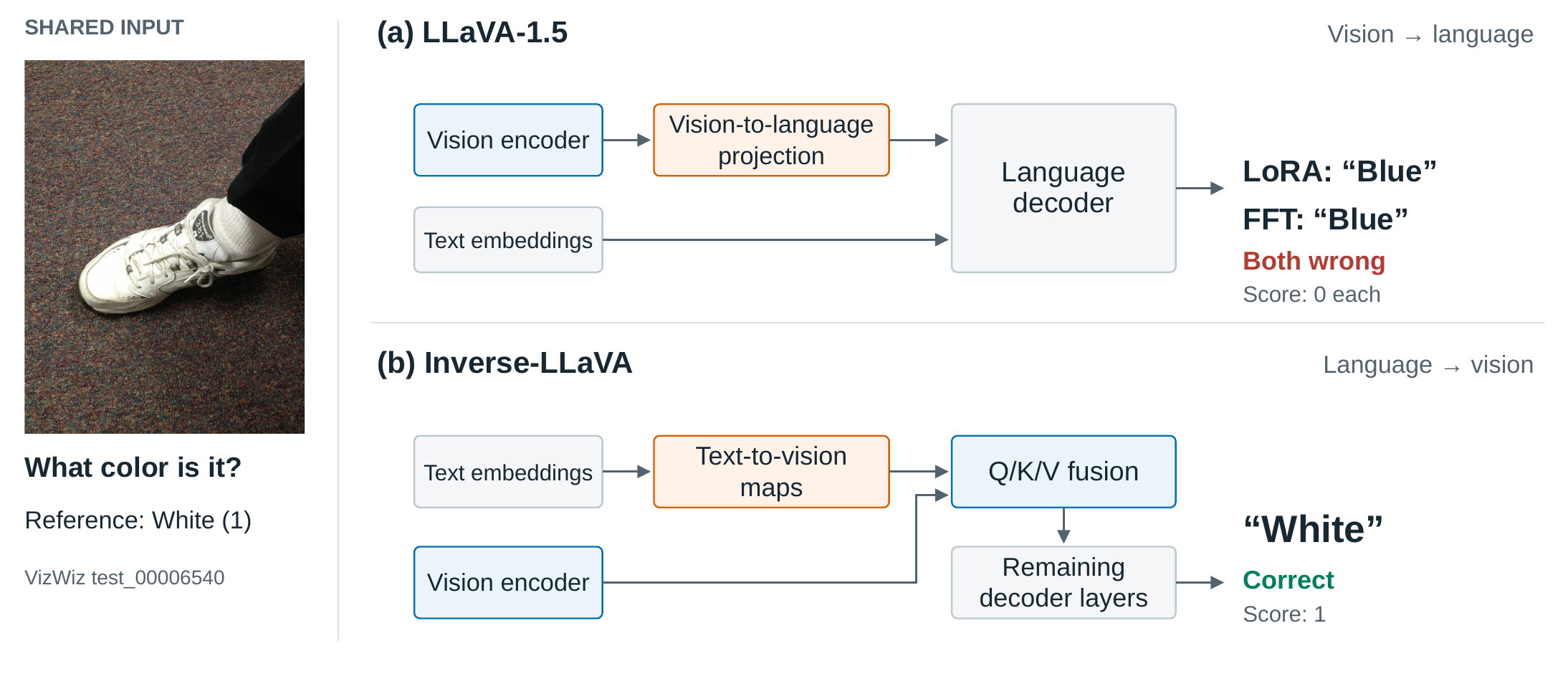}
\caption{\textbf{Inverse mapping for multimodal learning.} (a) LLaVA projects visual features to the language dimension before decoder processing. (b) Inverse-LLaVA maps language states to the visual dimension for attention fusion. Orange marks the cross-modal maps. The shared input is VizWiz \texttt{test\_00006540}; answers are verbatim from the three primary evaluations. The reference label gives the answer and consensus score (1, full credit). Inverse-LLaVA scores 1; both LLaVA references score 0. The fixed chat wrapper and response instructions are omitted from the display only. This selected example illustrates behavior, rather than its frequency. Figure~\ref{fig:structure} details the fusion branches and preserved language path.}
\label{fig:overview}
\end{figure*}

\begin{figure*}[!ht]
\centering
\includegraphics[width=\textwidth]{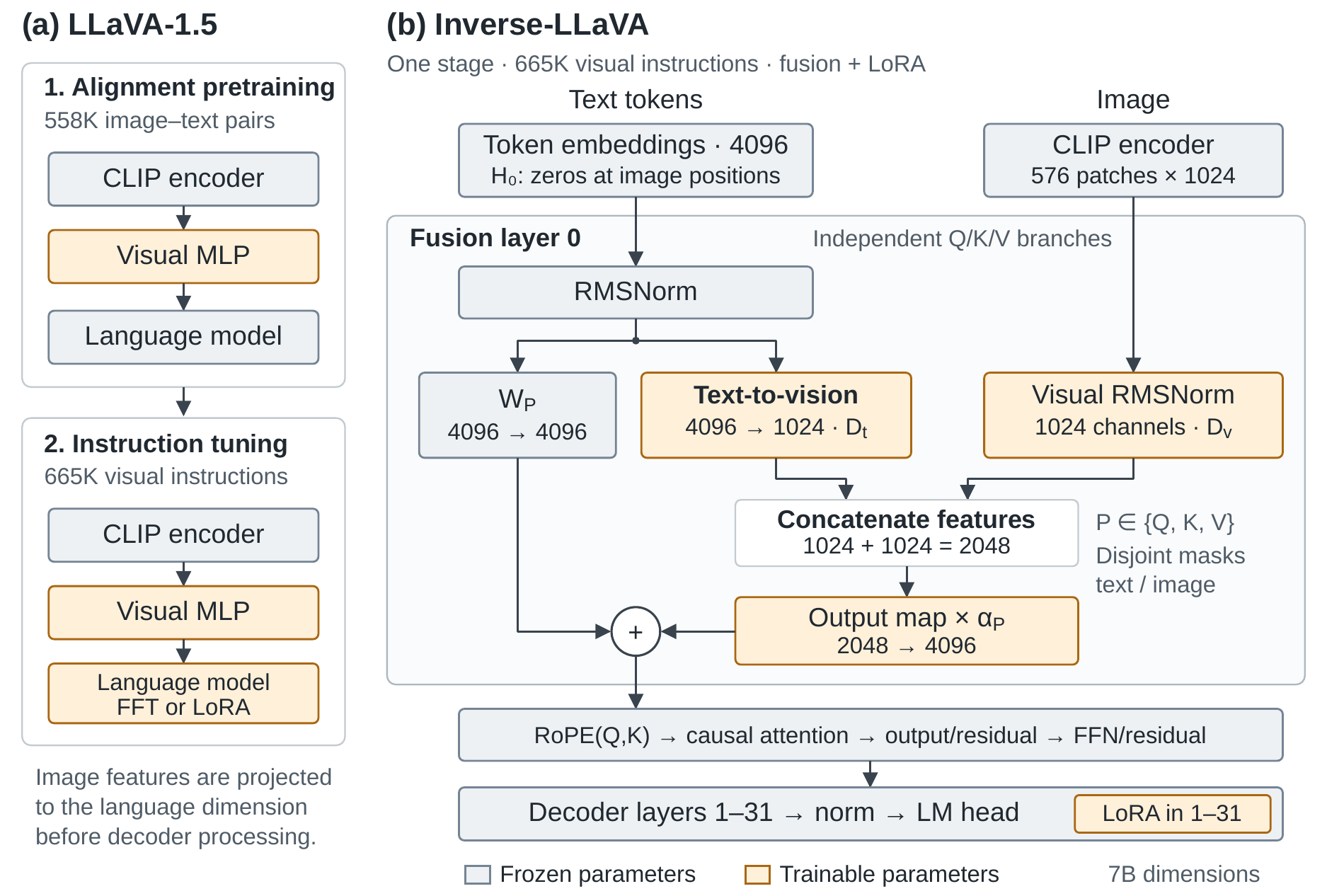}
\caption{\textbf{Architecture and training.} (a) LLaVA-1.5's visual-input pathway uses projector alignment before visual-instruction tuning; projected image features enter the decoder alongside text tokens. (b) Inverse-LLaVA jointly trains fusion and LoRA on instructions. The expanded sequence uses disjoint text/image masks $\mathbf{D}_t,\mathbf{D}_v$. Each Q/K/V branch has independent linear maps, visual normalization, and a scale. Feature concatenation preserves sequence length. The default 7B model inserts fusion at layer 0 and retains all 32 decoder layers; LoRA adapts layers 1--31. Equations~\eqref{eq:fusion_features}--\eqref{value_fusion} specify the updates.}
\label{fig:structure}
\end{figure*}

Inverse-LLaVA implements this idea through separate text-to-vision and fusion projections for queries, keys, and values. Encoder-width visual features enter these branches directly; learned updates then return to the required attention width. The resulting model jointly learns fusion and LoRA from visual instructions, without dedicated projector-alignment pretraining. Figure~\ref{fig:overview} illustrates the change in mapping direction, and Figure~\ref{fig:structure} shows the architecture and training recipes.

An instruction-only recipe makes the approach accessible using existing visual-instruction data. The final model reaches a similar performance range to the two-stage LLaVA-1.5 references on several benchmarks while omitting their 558K-example alignment stage. Supervision coverage remains an important experimental question: successful instruction training need not supply every kind of visual knowledge or answer behavior. Our continuation studies therefore hold the architecture fixed and vary paired and instruction supervision. They identify improved celebrity recognition, caption-induced response changes, and their repair through instruction replay.

\begin{samepage}
Our contributions are:
\begin{itemize}
\renewcommand{\labelitemi}{\(\bullet\)}
\item A text-to-vision attention interface that brings language states to the visual feature width and learns separate modality updates for queries, keys, and values.
\item A joint fusion--LoRA training recipe that connects frozen backbones using 665K visual instructions, reaching a similar performance range to two-stage LLaVA-1.5 on several benchmarks without its dedicated alignment stage.
\item An empirical characterization of the interface's computation, feature geometry, and supervision sensitivity, including controlled component and model-size studies.
\end{itemize}
\end{samepage}

\FloatBarrier
\section{Preliminaries and Related Work}
\label{sec:related}

\subsection{Multimodal Interfaces and Training Regimes}

LLaVA-1.5~\citep{liu2023improvedllava} first trains a visual projector using a 558K image--text mixture and then performs instruction tuning on a 665K mixture. The stages differ in both objectives and trainable components.

Single-stage training is an established design axis. Prismatic VLMs~\citep{karamcheti2024prismatic} examine training stages, visual backbones, image processing, and language models under controlled evaluations. This work makes a single-stage conventional projector an important reference for separating our fusion mechanism from the removal of an alignment stage.

Architectures also differ in where modalities interact. Flamingo uses cross-attention~\citep{alayrac2022flamingovisuallanguagemodel}; BLIP-2 uses a query transformer~\citep{li2023blip2bootstrappinglanguageimagepretraining}; CogVLM introduces visual expert modules~\citep{wang2023cogvlm}; and mPLUG-Owl2 studies modality-adaptive representations~\citep{ye2023mplug}. Inverse-LLaVA introduces text-to-vision fusion branches directly in selected decoder Q/K/V projections. Cambrian-1 studies visual encoders, connectors, and data-source balance~\citep{tong2024cambrian}, while Qwen3-VL incorporates multi-level visual features through DeepStack~\citep{bai2025qwen3vl}. Qwen2-VL~\citep{wang2024qwen2vlenhancingvisionlanguagemodels} and SEED-1.5-VL~\citep{guo2025seed15vltechnicalreport} further broaden the range of visual processing and training regimes.

InstructBLIP~\citep{dai2023instructblipgeneralpurposevisionlanguagemodels} conditions a query transformer on instructions while freezing the vision and language backbones. InternVL~\citep{chen2024internvl} scales visual pretraining and connects an InternViT encoder to the language model. EVE~\citep{diao2024EVE} removes the external vision encoder at inference, learning patch representations with visual-teacher supervision during training. Inverse-LLaVA instead retains frozen encoder features and learns text-to-vision attention updates jointly with LoRA. Table~\ref{tab:published_original_context} retains these comparisons, with their different training-data scopes in Table~\ref{tab:training_data_context}.

Later models also change supervision and visual processing. LLaVA-NeXT adds high-resolution crops and document/chart instruction data~\citep{liu2024llavanext}. Qwen2.5-VL~\citep{qwen2025qwen25vl} combines a dynamic-resolution vision encoder with spatial patch merging and a 4.1T-token training corpus; Qwen3-VL incorporates multi-level features through DeepStack. Tables~\ref{tab:published_context} and~\ref{tab:contemporary_reference} compare these newer systems using published results and our common-task Qwen3-VL evaluation, respectively. Appendix~\ref{app:contemporary_reference} details their backbones and supervision.

Vision as LoRA (VoRA)~\citep{wang2025vision} incorporates visual encoding into language-model adapters and transfers visual priors through block-wise distillation. Inverse-LLaVA retains an external frozen CLIP encoder and introduces separate text-to-vision and visual-feature branches within decoder attention. These distinctions separate the proposed interface from both conventional projectors and encoder-free adaptation.

\begin{samepage}
\subsection{Mapping Direction and Feature Width}

Both conventional visual projectors and inverse mappers operate on continuous hidden features. Mapping direction, modality-specific parameterization, and the location of interaction define the architectural distinction.
\par
\end{samepage}

Let an image encoder produce $p$ patch features in $\mathbb{R}^{d_v}$, and let language states have width $d_h$. A conventional projector maps $d_v$ to $d_h$ before decoder processing. Inverse-LLaVA maps language states from $d_h$ to $d_v$ within a fusion branch and then projects the fused features to the required attention width. For Vicuna-7B and CLIP ViT-L/14-336, $d_h=4096$ and $d_v=1024$. The inverse mapper is therefore a dimensionality reduction and is not generally injective.

\section{Methodology}
\label{sec:methodology}

Inverse-LLaVA freezes the CLIP encoder and Vicuna backbone, adds fusion to selected decoder attention layers, and learns LoRA adapters in the remaining layers. Fusion defaults to layer 0 (zero-based indexing), with all decoder layers retained.

\subsection{Image and Text Representations}

For one image and $n$ text tokens, expand the image placeholder into $p$ patch positions, giving a sequence of length $s=n+p$ before truncation. Let $\mathbf{H}_0\in\mathbb{R}^{d_h\times s}$ contain text embeddings at text positions and zeros at image positions. Let $\mathbf{V}_{\rm pad}\in\mathbb{R}^{d_v\times s}$ contain encoder features at image positions and zeros elsewhere. Diagonal masks $\mathbf{D}_t,\mathbf{D}_v\in\mathbb{R}^{s\times s}$ identify disjoint text and image positions. Their sum is the identity on valid positions; padding is excluded by the attention mask.

CLIP ViT-L/14-336 provides $p=576$ patch features. Inverse-LLaVA selects the final CLIP layer, with $d_v=1024$. Inverse-LLaVA-HD concatenates penultimate and final layer features along their channels, giving $d_v=2048$ at the same image resolution and patch count. Its mapper output width is therefore 2048 and its concatenated fusion branch is 4096 wide. HD tests compatibility with higher-dimensional visual features; it also changes feature content and parameter count, so this comparison alone does not isolate sensitivity to width.

\subsection{Text-to-Vision Q/K/V Fusion}

Let $\mathbf{H}_l$ be the residual stream entering decoder layer $l$ and $\widetilde{\mathbf{H}}_l=\operatorname{RMSNorm}_l(\mathbf{H}_l)$ its attention input, using root mean square layer normalization. For each branch $P\in\{Q,K,V\}$ at a selected layer, define
\begin{equation}
\mathbf{F}_l^P =
\operatorname{concat}_{\rm feat}\!\left(
\mathbf{W}_{\rm t2v}^{P,l}\widetilde{\mathbf{H}}_l\mathbf{D}_t,\,
\operatorname{RMSNorm}_{v}^{P,l}(\mathbf{V}_{\rm pad})\mathbf{D}_v
\right).
\label{eq:fusion_features}
\end{equation}
Here $\mathbf{W}_{\rm t2v}^{P,l}\in\mathbb{R}^{d_v\times d_h}$ is a branch-specific mapper, and $\mathbf{F}_l^P\in\mathbb{R}^{2d_v\times s}$. Concatenation is along feature width, so it does not double sequence length. The modality masks remain fixed as decoder states evolve.

The attention projections become
\begin{align}
\mathbf{Q}_l &= \mathbf{W}_Q^l\widetilde{\mathbf{H}}_l
 + \alpha_Q^l\mathbf{W}_{\rm up}^{Q,l}\mathbf{F}_l^Q,
\label{query_fusion}\\
\mathbf{K}_l &= \mathbf{W}_K^l\widetilde{\mathbf{H}}_l
 + \alpha_K^l\mathbf{W}_{\rm up}^{K,l}\mathbf{F}_l^K,
\label{key_fusion}\\
\mathbf{U}_l &= \mathbf{W}_V^l\widetilde{\mathbf{H}}_l
 + \alpha_V^l\mathbf{W}_{\rm up}^{V,l}\mathbf{F}_l^V.
\label{value_fusion}
\end{align}
We use $\mathbf{U}_l$ for attention values to distinguish them from visual encoder features. For Vicuna-7B, $\mathbf{W}_{\rm up}^{P,l}\in\mathbb{R}^{d_h\times2d_v}$. The pretrained $\mathbf{W}_P^l$ remain frozen in fusion layers. Each branch learns its mapper, visual RMSNorm gain, output projection, and scalar $\alpha_P^l$; the scalar is initialized to one. The mapper and output projections have no biases, consistent with the bias-free Vicuna attention projections; they contain no activation between mapping and concatenation.

Rotary positional embeddings (RoPE) are applied to queries and keys. Causal attention, output projection, residual addition, and the normalized feed-forward network (FFN) then follow the pretrained decoder structure.

\subsection{Optimization and Forward Computation}

Training minimizes autoregressive cross-entropy on assistant response tokens:
\begin{equation}
\mathcal{L}(\theta) =
-\frac{1}{|\mathcal{A}|}\sum_{i\in\mathcal{A}}
\log p_\theta(t_i\mid\mathcal{I},t_{<i}),
\end{equation}
where $\mathcal{A}$ contains supervised positions after prompt masking, patch expansion, and truncation. Image slots, user/system text, and padding are ignored in the loss.

\begin{samepage}
In the implementation, $\mathcal{A}$ is the set of supervised positions in each local microbatch. Gradient accumulation averages these microbatch losses; the logged token-weighted loss is a separate summary. Thus, equal global batch sizes alone do not guarantee identical token weighting across different microbatch partitions.
\par
\end{samepage}

The trainable set comprises all fusion parameters and LoRA matrices~\citep{hu2021lora}. LoRA uses rank $r=128$, alpha $\alpha_{\rm LoRA}=256$ (update multiplier $\alpha_{\rm LoRA}/r=2$), and dropout 0.05. Its adapters target attention and FFN projections outside fusion layers. The vision encoder, token embeddings, LM head, and underlying decoder weights remain frozen; the adapters and fusion branches modify their resulting computation.

\begin{algorithm}[!t]
\footnotesize
\caption{Inverse-LLaVA forward pass}
\label{alg:inverse_llava_forward}
\begin{algorithmic}[1]
\Require Image $\mathcal{I}$, text $\mathcal{T}$, fusion layers $\mathcal{S}\subseteq\{0,\ldots,L-1\}$
\Ensure Vocabulary logits at expanded sequence positions
\State Extract frozen visual patch features
\State Expand image placeholders and construct $\mathbf{H}_0,\mathbf{V}_{\rm pad},\mathbf{D}_t,\mathbf{D}_v$
\State Apply the sequence-length limit; construct valid-position and causal masks
\For{$l=0,\ldots,L-1$}
  \State $\widetilde{\mathbf{H}}_l\gets\operatorname{RMSNorm}_l(\mathbf{H}_l)$
  \If{$l\in\mathcal{S}$}
    \State Compute $\mathbf{Q}_l,\mathbf{K}_l,\mathbf{U}_l$ using Eqs.~\eqref{eq:fusion_features}--\eqref{value_fusion}
  \Else
    \State Compute the ordinary attention projections, including trainable LoRA updates
  \EndIf
  \State Apply RoPE to $\mathbf{Q}_l,\mathbf{K}_l$
  \State $\mathbf{R}_l\gets\mathbf{H}_l+\mathbf{W}_O^l\operatorname{CausalAttention}(\mathbf{Q}_l,\mathbf{K}_l,\mathbf{U}_l)$
  \State $\mathbf{H}_{l+1}\gets\mathbf{R}_l+\operatorname{FFN}_l(\operatorname{RMSNorm}'_l(\mathbf{R}_l))$
\EndFor
\State \Return $\mathbf{W}_{\rm lm}\operatorname{RMSNorm}_{\rm final}(\mathbf{H}_L)$
\end{algorithmic}
\end{algorithm}

\subsection{Complexity and Design Choices}
\label{sec:complexity}

For one fusion layer with three full-width branches, the dense mapper/output projections require $9sd_hd_v$ multiply--accumulates per prefill, excluding normalization and scalar operations. They add $9d_hd_v+3d_v+3$ parameters: 37,751,811 for the 7B dimensions. Decoder attention still processes the expanded sequence, with its usual quadratic prefill attention term. Cached generation avoids recomputing prior keys and values.

LLaVA-1.5's two-layer visual MLP has $d_vd_h+d_h^2+2d_h$ parameters (20,979,712 here) and requires $p(d_vd_h+d_h^2)$ projection multiply--accumulates per image. It is evaluated once before decoder processing; inverse fusion also operates on subsequent text positions. Counting a multiply--accumulate as two FLOPs, these expressions compare interface work only, excluding backbone computation and elementwise operations. Removing a training stage is a separate source of potential savings.

At fixed $d_v=1024$, increasing the language width from Vicuna-7B's 4096 to Vicuna-13B's 5120 gives 47,188,995 fusion parameters and 25\% more fusion-projection work per token. This linear dependence on language width describes interface cost; the larger decoder contributes additional computation. It does not predict accuracy scaling.

Addition shares one modality-output map; concatenation keeps two. Gating also changes parameterization, so these comparisons involve capacity and optimization.

For one branch, omit layer and target indices and partition $\mathbf{W}_{\rm up}=[\mathbf{B}_t\;\mathbf{B}_v]$, giving
\[
\Delta\mathbf{P}=\alpha\bigl(
\mathbf{B}_t\mathbf{W}_{\rm t2v}\widetilde{\mathbf{H}}\mathbf{D}_t
+\mathbf{B}_v\operatorname{RMSNorm}_v(\mathbf{V}_{\rm pad})\mathbf{D}_v\bigr).
\]
At text positions, the update matrix $\alpha\mathbf{B}_t\mathbf{W}_{\rm t2v}$ has rank at most $d_v$; image positions use a separate visual output map. Cross-token interaction occurs in attention. Matching the visual feature width specifies an interface, without imposing semantic alignment of the mapped text coordinates. The learned scalar can be absorbed into the output map at inference, so the fixed-scale comparison tests optimization parameterization rather than additional representational capacity.

We use the first layer so that visual features participate from the beginning of decoder processing. The formulation also permits later insertion and higher-dimensional visual features. Section~\ref{sec:components} tests these choices at full decoder depth.

\section{Experiments}

\subsection{Experimental Setup}
\label{sec:setup}

We use Vicuna-7B-v1.5~\citep{zheng2023judgingllmasajudgemtbenchchatbot} and CLIP ViT-L/14-336~\citep{radford2021learningtransferablevisualmodels}, the same pretrained backbone families as LLaVA-1.5. Training uses two H100 SXM 80 GB GPUs with effective batch 128: 32 examples per GPU and two accumulation steps. Execution uses bfloat16 (BF16), gradient checkpointing, scaled dot-product attention (SDPA), and DeepSpeed ZeRO-2, with a 2,048-position expanded context. The final checkpoint at update 5,198 is used throughout the primary evaluation; it is not selected by benchmark performance.

\noindent\textbf{Training Configuration.} Inverse-LLaVA uses the LLaVA-v1.5-mix665k visual-instruction mixture for one epoch, with learning rate $2\times10^{-4}$, cosine scheduling, weight decay 0, and warmup ratio 0.03. The prepared dataset contains 665,298 rows, including text-only examples. The data manifest records source revisions, OCR-VQA image provenance, and image-integrity audits. LoRA settings for our model are given in Section~\ref{sec:methodology}.

The logged training intervals retained in the final checkpoint total 19.75 hours on two H100s (39.49 GPU-hours). Peak allocated and reserved memory are 60.26 and 74.44 GiB per GPU, respectively, taking the maximum across ranks. These measurements describe our run; a matched LLaVA training-time comparison is not available. Appendix~\ref{app:reproduction_protocols} distinguishes retained training time from additional execution overhead.

\noindent\textbf{Baseline identities.} We re-evaluate the official LLaVA-1.5-7B LoRA and full-fine-tuning (FFT) checkpoints using the same benchmark inputs and scoring rules as Inverse-LLaVA (Table~\ref{tab:original_nine} and Figure~\ref{fig:canonical_mme}). LoRA provides the parameter-efficient baseline and FFT the fully fine-tuned baseline. Both use penultimate CLIP features; their alignment data and adapter targets also differ from ours. Table~\ref{tab:model_context} specifies the architecture and training differences. The matched single-stage study in Appendix~\ref{app:matched_interface} separately controls supervision and feature selection.

\vspace{-0.5ex}
\begin{samepage}
\noindent\textbf{Evaluation Protocol.} Table~\ref{tab:original_nine} specifies MM-Vet~\citep{yu2023mmvet}, MMBench EN/CN~\citep{liu2023mmbench}, MME perception~\citep{fu2023mme}, VizWiz~\citep{gurari2018vizwiz}, ScienceQA-IMG~\citep{lu2022scienceqa}, VQAv2~\citep{yash2017vqa2}, TextVQA~\citep{singh2019textvqa}, and GQA~\citep{hudson2019gqa}. MME cognition provides a category-level reasoning analysis; OCRBench~\citep{liu2024ocrbench}, AI2D~\citep{kembhavi2016diagram}, and MMStar~\citep{chen2024mmstar} extend task coverage. Appendix~\ref{app:reproduction_protocols} specifies splits, prompts, generation limits, and scoring rules. Predictions, input hashes, and evaluator identities are retained for every reported result.
\par
\end{samepage}

\subsection{Benchmark Results}
\label{sec:canonical_evaluation}

Table~\ref{tab:original_nine} reports the nine primary benchmark evaluations, separating published LLaVA-1.5 scores from our common-protocol evaluations of the official checkpoints. Appendix Table~\ref{tab:canonical_panel} adds supplementary tasks and MME cognition. The broader LLaVA-1.5 suite also includes POPE, SEED-Bench, and LLaVA-Bench-in-the-Wild.

Unless stated otherwise, comparisons use 95\% paired bootstrap intervals conditional on the fitted checkpoints. These quantify benchmark-sampling uncertainty, exclude training-seed variation, and are unadjusted for multiple comparisons. An interval containing zero does not establish equivalence. Table emphasis ranks point estimates and does not indicate statistical significance. Appendix~\ref{app:reproduction_protocols} details the bootstrap procedure.

\begin{table}[!ht]
\centering
\small
\caption{\textbf{Primary benchmark results.} Published LLaVA-1.5 scores~\citep{llavamodelzoo} are separated from evaluations of the final Inverse-LLaVA and official LoRA/FFT checkpoints under common task protocols. Values are percentages except MME. VQAv2 uses official test-dev server scores; MM-Vet uses the hosted GPT-4.1 judge. Full protocols appear in Table~\ref{tab:local_protocols}.}
\label{tab:original_nine}
\setlength{\tabcolsep}{4pt}
\begin{tabular}{@{}lrr@{\hspace{12pt}}>{\columncolor{methodshade}}rrr@{}}
\toprule
 & \multicolumn{2}{c}{Published LLaVA-1.5} & \multicolumn{3}{c}{Re-evaluated, common protocols} \\
\cmidrule(lr){2-3}\cmidrule(l){4-6}
Benchmark & LoRA & FFT & Inverse-LLaVA & LoRA & FFT \\
\midrule
MM-Vet v1 & 30.2 & 31.1 & 28.67 & \textbf{31.24} & \underline{29.68} \\
VizWiz test & 47.8 & 50.0 & \textbf{50.96} & 48.56 & \underline{50.64} \\
ScienceQA-IMG & 68.4 & 66.8 & \textbf{69.61} & 68.82 & \underline{69.11} \\
MMBench EN dev & 66.1 & 64.3 & 62.63 & \textbf{67.10} & \underline{65.12} \\
MMBench CN dev & 58.9 & 58.3 & 54.04 & \textbf{58.93} & \underline{58.33} \\
MME perception & 1476.9 & 1510.7 & 1453.82 & \underline{1484.58} & \textbf{1507.28} \\
VQAv2 test-dev & 79.1 & 78.5 & 78.45 & \textbf{79.13} & \underline{78.55} \\
TextVQA val & 58.2 & 58.2 & 56.96 & \textbf{58.47} & \underline{58.21} \\
GQA balanced test-dev & 63.0 & 62.0 & \underline{62.28} & \textbf{62.63} & 61.89 \\
\bottomrule
\end{tabular}
\par\smallskip\begin{minipage}{0.98\linewidth}\footnotesize Re-evaluated row best/second: bold/underline. Gray: Inverse-LLaVA.\end{minipage}
\end{table}

\noindent\textbf{Performance with an instruction-only recipe.} Without the baselines' 558K-example alignment stage, Inverse-LLaVA reaches 78.45\% on VQAv2 versus LoRA's 79.13\%, and 62.28\% on GQA versus 62.63\%. VizWiz improves by 2.40 percentage points ($[1.58,3.21]$), with an answerability analysis locating the gain in unanswerable questions. Its 0.32-point difference from FFT has an interval containing zero. MMBench EN/CN are lower than LoRA by 4.47/4.90 points, with both intervals below zero. All EN answers parse, and correcting every unparsed CN answer could recover at most 0.60 points. These deficits therefore extend beyond answer extraction.

\begin{figure*}[!ht]
    \centering
\includegraphics[width=0.99\textwidth]{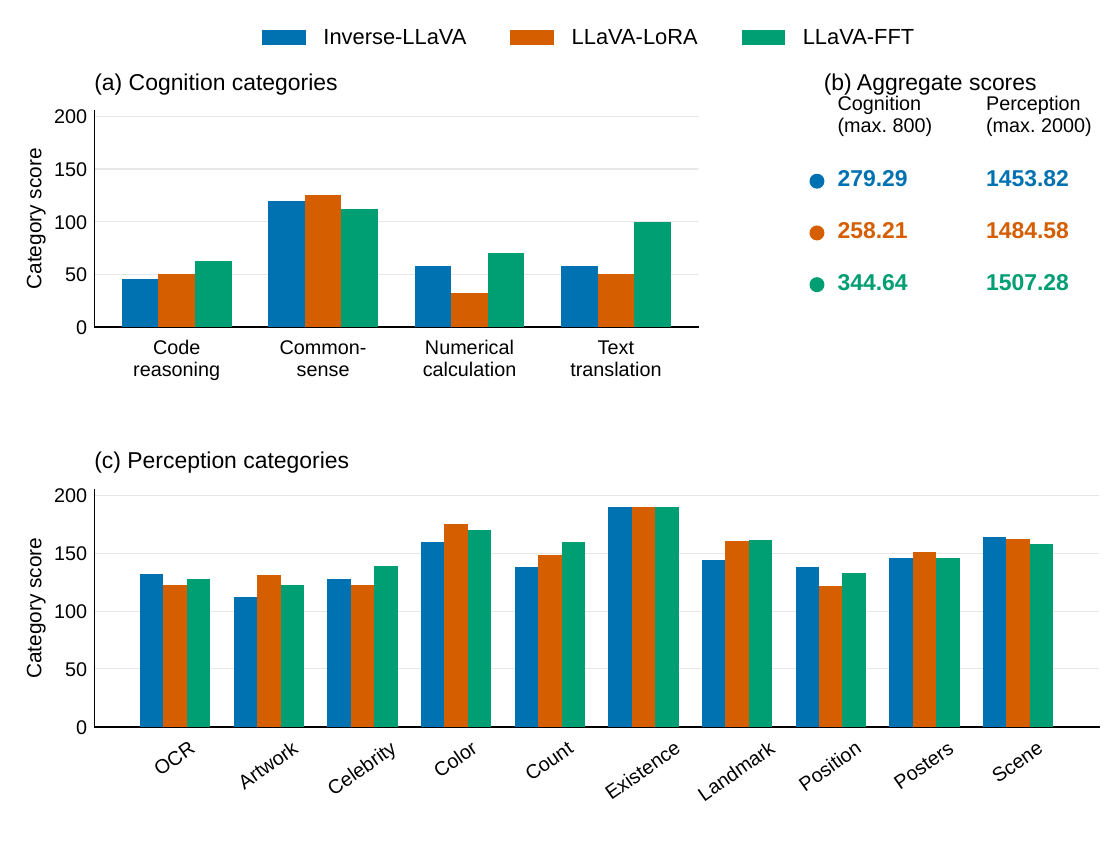}
    \caption{\textbf{Where the MME differences occur.} The same final checkpoints, questions, and official scoring rules are used throughout. (a) Cognition categories, (b) aggregate scores, and (c) perception categories retain their native units. Inverse-LLaVA's cognition point score exceeds LoRA's but falls below FFT's; the category breakdown shows that this is not a uniform reasoning advantage. Each category has a maximum of 200. Aggregate paired intervals are reported in the text.}
    \label{fig:canonical_mme}
\end{figure*}

\noindent\textbf{MME categories.} Figure~\ref{fig:canonical_mme} covers all 260 cognition and 2,114 perception questions, using common prompts and the pinned official scorer. Category scores reveal which tasks contribute to the aggregate differences.

Inverse-LLaVA obtains cognition 279.29 and perception 1453.82. The corresponding LoRA scores are 258.21 and 1484.58; FFT scores are 344.64 and 1507.28. Using 10,000 paired, image-grouped bootstrap resamples with category-aware aggregation, the cognition difference against LoRA is $+21.07$ with a 95\% interval of $[-20.36,63.93]$. The perception difference is $-30.76$ with interval $[-85.66,23.75]$. Both intervals span zero.

Inverse-LLaVA scores 132.50 on MME OCR, compared with LoRA 122.50 and FFT 127.50; its celebrity score is 127.65. These category-level point differences are distinct from TextVQA and OCRBench outcomes. The original MME calculator independently reproduces all three models' category scores from saved predictions.

\noindent\textbf{TextVQA validation.} On all 5,000 questions with the LLaVA prompts and Rosetta OCR inputs, Inverse-LLaVA scores 56.96\%, versus LoRA 58.47\% and FFT 58.21\%. The paired differences are $-1.51$ percentage points ($[-2.33,-0.65]$) and $-1.25$ ($[-2.12,-0.38]$). This deficit motivates our supervision interventions.

\noindent\textbf{Reasoning evaluations.} With released ScienceQA CQM-A prompts, Inverse-LLaVA reaches 69.61\%, versus 68.82\% for LoRA and 69.11\% for FFT. The differences are $+0.79$ points ($[-0.70,2.30]$) and $+0.50$ ($[-1.12,2.13]$). Together with mixed AI2D and MMStar results, these estimates support task-specific comparisons rather than a general reasoning advantage. Appendix~\ref{app:benchmark_differences} reports the remaining benchmark differences and response analyses; VQAv2 provides only server-level aggregate scores.

\noindent\textbf{Broader model comparison.} Tables~\ref{tab:published_original_context} and~\ref{tab:published_context} place these results alongside InstructBLIP, InternVL-Chat, EVE, LLaVA-1.6, and the Qwen-VL family. Newer systems substantially raise scene-text and multimodal benchmark scores: Qwen2.5-VL-7B reports 84.9\% TextVQA and Qwen3-VL-8B reports 896/1000 on OCRBench. Their training and visual processing differ from the frozen-CLIP, 665K-instruction setting examined here. Table~\ref{tab:training_data_context} separates backbone pretraining from connector alignment, and Appendix~\ref{app:contemporary_reference} details the newer recipes. These comparisons locate the instruction-only result within the broader field, while the matched studies below examine interface design at fixed instruction exposure.

\begin{table*}[!ht]
\centering
\small
\setlength{\tabcolsep}{6pt}
\caption{Published results for related Vicuna-7B systems~\citep{dai2023instructblipgeneralpurposevisionlanguagemodels,chen2024internvl,diao2024EVE,liu2023improvedllava}. Scores are percentages except MME perception; a dash denotes an unreported result. Architectures, training recipes, and evaluation protocols differ from the common-protocol comparison in Table~\ref{tab:original_nine}.}
\label{tab:published_original_context}
\begin{tabular}{lrrr}
\toprule
Benchmark & InstructBLIP & InternVL-Chat & EVE-7B \\
\midrule
MM-Vet & 26.2 & -- & 25.6 \\
VizWiz & 34.5 & 52.5 & 41.8 \\
ScienceQA-IMG & 60.5 & -- & 63.0 \\
MMBench EN & 36.0 & -- & 49.5 \\
MMBench CN & 23.7 & -- & -- \\
MME perception & -- & 1525.1 & 1217.3 \\
VQAv2 & -- & 79.3 & 75.4 \\
TextVQA & 50.1 & 57.0 & 51.9 \\
GQA & 49.2 & 62.9 & 60.8 \\
\bottomrule
\end{tabular}
\end{table*}
\begin{table*}[!ht]
\centering
\small
\caption{\textbf{Published results from newer VLMs.} Values follow the released model zoo and model cards~\citep{llavamodelzoo,qwen2vlcard,qwen25vlcard,qwen3vlcard}. Scores are percentages except the stated OCRBench and MME scales. Source-specific splits and judges occupy separate rows. Table~\ref{tab:model_context} gives architecture and training context; Table~\ref{tab:contemporary_reference} gives our common-task Qwen3-VL evaluation.}
\label{tab:published_context}
\setlength{\tabcolsep}{4pt}
\begin{tabular}{@{}lrrrr@{}}
\toprule
Benchmark / source protocol & \shortstack{LLaVA-1.6\\Vicuna-7B} & \shortstack{Qwen2-VL\\7B} & \shortstack{Qwen2.5-VL\\7B} & \shortstack{Qwen3-VL\\8B} \\
\midrule
TextVQA val & 64.9 & 84.3 & 84.9 & -- \\
OCRBench (out of 1,000) & -- & 845 & 864 & 896 \\
MMStar & -- & 60.7 & 63.9 & 70.9 \\
AI2D test & -- & -- & -- & 85.7 \\
\midrule
VQAv2 & 81.8 & -- & -- & -- \\
GQA & 64.2 & -- & -- & -- \\
VizWiz & 57.6 & -- & -- & -- \\
ScienceQA-IMG & 70.1 & -- & -- & -- \\
MME perception (out of 2,000) & 1519 & -- & -- & -- \\
MME total (out of 2,800) & -- & 2326.8 & -- & -- \\
\midrule
MMBench EN/CN (LLaVA release) & 67.4/60.6 & -- & -- & -- \\
MMBench EN/CN test & -- & 83.0/80.5 & -- & -- \\
MMBench v1.1 EN test & -- & 80.7 & 82.6 & -- \\
MMBench v1.1 EN dev & -- & -- & -- & 85.0 \\
MM-Vet (LLaVA release judge) & 43.9 & -- & -- & -- \\
MM-Vet (GPT-4-Turbo) & -- & 62.0 & 67.1 & -- \\
\bottomrule
\end{tabular}
\par\smallskip\begin{minipage}{0.98\linewidth}\footnotesize
LLaVA-1.6 uses FFT; the Qwen entries are Instruct models. A dash means no value is included from the cited source. Published rows are contextual and are not ranked with our common-protocol evaluations. The Qwen3-VL card recommends sampled generation; our separate evaluation uses greedy decoding.
\end{minipage}
\end{table*}
\begin{table*}[!ht]
\centering
\small
\caption{Training-data context for the compared Vicuna-7B systems. Counts describe the indicated pools or stages, rather than a common measure of total training cost. All models also inherit language pretraining. The CLIP-based systems inherit visual pretraining, including Inverse-LLaVA.}
\label{tab:training_data_context}
\begin{tabular}{lp{0.53\textwidth}r}
\toprule
Model & Visual prior and pre-instruction stages & Instructions \\
\midrule
LLaVA-1.5 & Pretrained CLIP; 558K alignment pairs & 665K \\
InstructBLIP & BLIP-2 initialization from 129M image--text pairs; frozen visual encoder and LLM & 1.2M \\
InternVL-Chat & InternViT: 4.98B cleaned pretraining pairs; this MLP variant uses 558K alignment examples & 665K \\
EVE-7B & Visual-teacher supervision; 16M then 33M caption exposures from a 33M pool & 665K \\
\rowcolor{methodshade}
Inverse-LLaVA & Pretrained CLIP; instruction tuning only & 665K \\
\bottomrule
\end{tabular}
\end{table*}

For a common-task comparison, we evaluate official Qwen3-VL-8B-Instruct on the same 1,164 MMBench EN dev groups and 4,329 rotations, retaining its native image processing and chat template. Table~\ref{tab:contemporary_reference} shows 84.97\% circular accuracy, compared with 62.63\% for Inverse-LLaVA, 67.10\% for LoRA, and 65.12\% for FFT. The corresponding paired Qwen3-VL gains are 22.34, 17.87, and 19.85 percentage points, with all intervals above zero. This measures system-level performance under common scoring; the models retain their different backbones and training histories.

\begin{table}[!ht]
\centering
\small
\caption{MMBench EN dev under common questions and circular scoring: 1,164 groups and 4,329 rotations. Qwen3-VL uses its native image processor and chat template. Backbones and training recipes differ, as detailed in Table~\ref{tab:model_context}.}
\label{tab:contemporary_reference}
\begin{tabular}{lrrr}
\toprule
Model & Circular (\%) & Base rotation (\%) & Unparsed \\
\midrule
\rowcolor{methodshade}
Inverse-LLaVA & 62.63 & 72.16 & 0 \\
LLaVA-LoRA & \underline{67.10} & \underline{74.91} & 0 \\
LLaVA-FFT & 65.12 & 74.48 & 0 \\
Qwen3-VL-8B-Instruct & \textbf{84.97} & \textbf{89.95} & 11 \\
\bottomrule
\end{tabular}
\par\smallskip\begin{minipage}{0.98\linewidth}\footnotesize Bold/underlined: best/second-best accuracy. Gray: Inverse-LLaVA.\end{minipage}
\end{table}

\subsection{Component Screening}
\label{sec:components}

Table~\ref{tab:component_screen} compares five conditions trained from pretrained Vicuna/CLIP with fresh adapters on the same 5\% instruction subset. All use seed 42, 260 updates, and global batch 128. Compatible initialization tensors are identical; differently shaped output maps follow the declared initialization. Frozen-mapper and fixed-scale variants change only trainability. Gating uses one sigmoid gate shared across Q/K/V, with zero-initialized weights and bias and a branch-specific output matrix applied to the gated sum. Appendix~\ref{app:reproduction_protocols} gives exposure and response audits.

\begin{table}[!ht]
\centering
\small
\caption{Component ablations on the same 5\% instruction subset: 260 updates, global batch 128, and shared compatible initialization. OCRBench and MME cognition use complete splits. Each condition has its own 260-update schedule, separate from the full-run trajectory.}
\label{tab:component_screen}
\begin{tabular}{lrrr}
\toprule
Condition & Trainable (M) & OCRBench (\%) & MME cognition \\
\midrule
\rowcolor{methodshade}
Concatenation reference & 347.57 & 2.30 & \underline{288.57} \\
Frozen text-to-vision mapper & 334.99 & \textbf{3.00} & 277.86 \\
Fixed fusion scales & 347.57 & \underline{2.70} & 285.71 \\
Addition & 334.99 & 2.20 & 242.14 \\
Gating & 337.09 & 2.20 & \textbf{303.57} \\
\bottomrule
\end{tabular}
\par\smallskip\begin{minipage}{0.98\linewidth}\footnotesize Bold/underlined: highest/second-highest score per metric. Gray marks the reference.\end{minipage}
\end{table}

Concatenation exceeds addition by 46.43 MME cognition points (95\% paired interval $[15.35,82.86]$) and 0.10 OCR percentage points ($[-0.60,0.72]$). Gating minus concatenation is 15.00 cognition points ($[-30.00,60.00]$) and $-0.10$ OCR percentage points ($[-0.89,0.69]$). Under this budget, concatenation improves cognition over addition; gating shows no clear degradation.

Near-floor OCR, all-``No'' responses within some MME categories, and unequal output-map capacities limit full-data interpretation (Appendix~\ref{app:reproduction_protocols}).

Fusion depth and CLIP-layer selection test different design choices. Moving fusion from layer 0 to layer 16, with matched LoRA placement and training data, gives MME cognition scores of 289.29 and 287.86 and identical OCRBench scores of 2.20\%. In a four-condition comparison, penultimate CLIP features improve ScienceQA-IMG for both inverse fusion and the early visual projector. The interaction interval contains zero, so the study does not establish a feature benefit specific to inverse fusion. TextVQA does not improve. Appendix~\ref{app:matched_interface} reports the paired intervals and controls.

The same formulation accommodates 2,048-dimensional visual features in Inverse-LLaVA-HD and a Vicuna-13B language backbone. Both matched 5\% studies evaluate three checkpoints (Appendices~\ref{app:matched_width} and~\ref{app:matched_language}). HD's final comparison intervals include zero. The 13B model has higher TextVQA and ScienceQA-IMG point scores at every checkpoint, although the final intervals also span zero. Its lower MME cognition score is concentrated in Chinese text translation; supplying literal transcriptions substantially improves both model sizes. Training cost rises from 2.15 to 3.36 GPU-hours. These experiments demonstrate compatibility with larger visual representations and language backbones and identify task-dependent scaling behavior. Their single-seed, limited-data setting does not establish a full-data scaling law.

\subsection{Computation and Inference}
\label{sec:diagnostics}

Inverse-LLaVA trains 347.57M parameters, versus 340.80M for official LLaVA-LoRA and 6.76B for FFT. The larger fusion module outweighs the saving from excluding LoRA at the fusion layer. Appendix~\ref{app:reproduction_protocols} gives exact total and trainable counts, including the FFT checkpoint's vocabulary-conversion difference.

\begin{figure*}[!t]
\centering
\includegraphics[width=\textwidth]{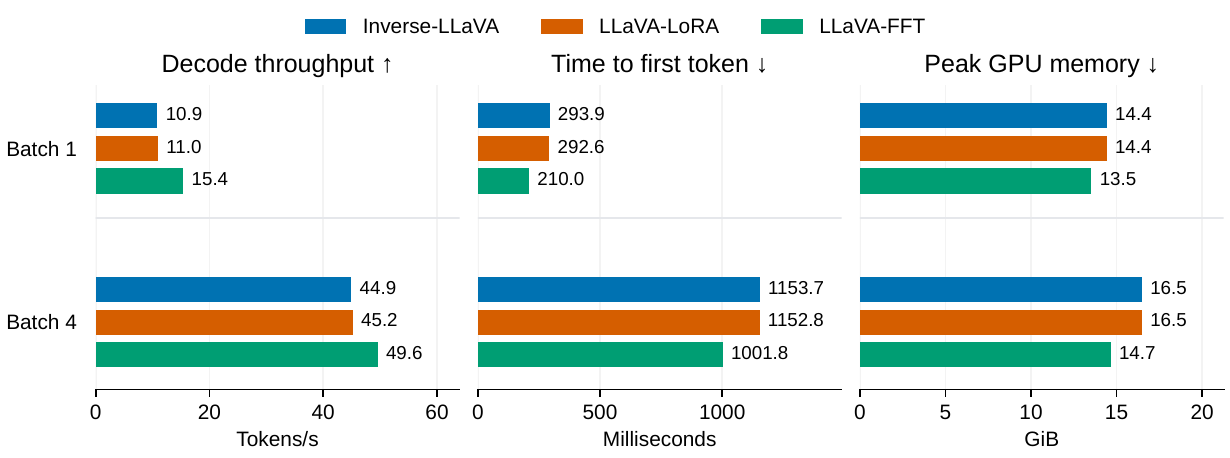}
\caption{Fixed-work inference profiles on NVIDIA Thor. Bars show median throughput and time to first token, and peak allocated GPU memory. Prefill and decoding exclude separately measured image/prompt preparation. All models use identical inputs and BF16/SDPA; the two adapter-based models use unmerged LoRA. The official FFT reference uses the Transformers LLaVA path.}
\label{fig:canonical_efficiency}
\end{figure*}

We profile all three models on one NVIDIA Thor with BF16/SDPA, identical prompts, batch sizes 1 and 4, ten warmups, and thirty measured repetitions. Decoding generates 32 tokens with EOS ignored to equalize work; LoRA remains unmerged. At batch size one, median throughput is 10.86 tokens/s for Inverse-LLaVA, 10.97 for LoRA, and 15.41 for FFT. Prefill-to-first-token medians are 293.9, 292.6, and 210.0 ms; image/prompt preparation is measured separately (1.038, 1.038, and 0.939 s). Figure~\ref{fig:canonical_efficiency} reports both batch sizes. These platform- and execution-specific timings characterize inference; they do not measure training savings or predict H100 latency.

The training recipe omits the 558K-example alignment stage and its projector-training work. Under a one-epoch convention, training on 665K rather than 558K + 665K examples reduces the number of multimodal training examples processed by approximately 45.6\%. This count excludes backbone pretraining, which both recipes inherit. Alignment and instruction tuning have different per-example costs, and inverse fusion adds the computation in Section~\ref{sec:complexity}; the sample reduction therefore cannot be interpreted as a 45.6\% time or FLOP saving.

\subsection{Representations and Text-Only Behavior}
\label{sec:representations}

We capture the last valid prompt-token state at every decoder output for one fixed OCRBench panel of 100 distinct questions and 100 distinct images. Linear centered kernel alignment (CKA)~\citep{kornblith2019similarity} between Inverse-LLaVA and LoRA rises from 0.782 at output 1 to 0.977 at output 16. Final-output similarities are 0.950 against LoRA (95\% paired-row interval $[0.946,0.967]$) and 0.885 against FFT ($[0.867,0.912]$). Appendix~\ref{app:canonical_geometry} gives the sampling rule, cross-layer matrices, blank-image controls, and estimator checks. The pattern locates early representation differences followed by stronger correspondence. Shared backbones and prompt formatting, alongside differing adapter and CLIP-layer choices, limit attribution to fusion alone.

\noindent\textbf{Similarity ordering and geometric spread.} Mapping text from 4,096 to 1,024 dimensions preserves much of the pairwise similarity ordering: representational similarity analysis (RSA), measured by Spearman correlation of cosine similarities, gives 0.803/0.810/0.831 for the Q/K/V branches. Dimension reduction has different geometric effects across branches: median pairwise cosine distance decreases to 0.040/0.013 times its input value for Q/K and increases to 4.163 times for V. The conventional visual projectors also retain substantial ordering (RSA 0.902 for LoRA and 0.839 for FFT). These measurements show that a smaller feature dimension need not imply uniform geometric contraction. Appendix~\ref{app:projection_geometry} combines the high-dimensional measurements with principal component analysis (PCA) and t-distributed stochastic neighbor embedding (t-SNE) views; the inverse measurements include normalization and mapping.

The qualitative examples in Appendix~\ref{app:benchmark_cases} span visual questions, scientific concepts, binary text verification, and open-ended scene-text answers. The same inputs are shown with responses from the final Inverse-LLaVA checkpoint and both official references. Outcome-stratified selection includes successes and failures; full prompts, including OCR hints, and unedited answers accompany each example.

\noindent\textbf{Text-only evaluation.} Appendix~\ref{app:language_retention} reports HellaSwag scores of 73.18\% for Inverse-LLaVA, 71.44\% for LoRA, and 71.58\% for Vicuna. Paired gains are 1.74 points ($[1.21,2.28]$) and 1.60 ($[1.03,2.18]$). FFT reaches 74.46\%, exceeding Inverse-LLaVA by 1.27 points ($[0.80,1.73]$). All ARC-Challenge comparison intervals include zero. The HellaSwag results demonstrate useful text-only performance after multimodal adaptation; these two tasks do not establish retention of every language capability.

\subsection{Supervision Interventions}
\label{sec:supervision}

The instruction-only result motivates examining which behaviors change when the same model receives additional visual supervision. Inverse-LLaVA consumes caption pairs through its autoregressive pipeline while jointly adapting LoRA and fusion. Its instruction mixture already contains book-oriented OCR-VQA and TextCaps captioning examples~\citep{liu2023improvedllava}. TextCaps uses the TextVQA image directory and OCR hints with caption targets, so the mixture's OCR-related exposure covers different objectives from scene-text question answering.

We independently continue the final checkpoint on 1\% or 5\% paired data, OCR-targeted instruction examples, and instruction/non-OCR controls. These continuations hold the architecture fixed and vary supervision content and objective. Table~\ref{tab:recovery_exposure} records their row counts, optimizer updates, and supervised tokens.

MME celebrity recognition increases from 127.65 for the parent checkpoint to 152.94 after paired 1\% continuation, a gain of 25.29 points (95\% paired-image interval $[15.88,35.29]$). Fully correct celebrity image pairs increase to 117 of 170. The gain relative to the instruction-token control is 32.65 points ($[22.06,43.53]$). Aggregate perception changes are less favorable: paired 1\% changes the total by 20.15 points ($[-20.90,61.49]$), while paired 5\% changes it by $-87.80$ ($[-142.92,-32.01]$). Appendix~\ref{app:recovery_categories} retains every perception category and control. The tested supervision improves celebrity recognition, without broad perception recovery.

Table~\ref{tab:recovery_panel} reports all continuation conditions. Caption-only continuations score 54.25\% on TextVQA for paired 1\% and 46.82\% for paired 5\%, compared with the parent's 56.96\%. Their differences from the parent are $-2.71$ percentage points ($[-3.50,-1.93]$) and $-10.14$ points ($[-11.23,-9.03]$). OCR replay and all three instruction controls remain within 0.29 points of the parent, with intervals containing zero; their OCRBench intervals also contain zero. The contrasting caption and instruction results motivate examining the adaptation objective and response behavior.

\begin{table*}[!ht]
\centering
\small
\caption{Supervision-continuation evaluations. Accuracy metrics are percentages; MME uses score points. All continuations start independently from the final Inverse-LLaVA checkpoint. Token controls are approximate and have different update counts. ScienceQA uses the official case-sensitive extractor; response-format diagnostics are reported separately in the text.}
\label{tab:recovery_panel}
\begin{tabular}{lrrrrr}
\toprule
Condition & OCRBench & TextVQA & MME cog. & SQA-IMG & MME per. \\
\midrule
\rowcolor{methodshade}
Inverse-LLaVA & 31.00 & 56.96 & 279.29 & 69.61 & 1453.82 \\
Paired 1\% & 29.30 & 54.25 & 321.43 & 54.14 & 1473.97 \\
Paired 5\% & 25.30 & 46.82 & 324.64 & 2.28 & 1366.03 \\
OCR replay & 31.00 & 56.82 & 280.00 & 69.76 & 1432.55 \\
Instruction (tokens) & 31.40 & 56.67 & 282.14 & 69.31 & 1410.97 \\
Non-OCR (rows) & 30.90 & 56.99 & 276.43 & 69.46 & 1443.99 \\
Non-OCR (tokens) & 30.70 & 56.98 & 290.00 & 69.41 & 1434.63 \\
\bottomrule
\end{tabular}
\par\smallskip\begin{minipage}{0.98\linewidth}\footnotesize Gray marks the unchanged Inverse-LLaVA parent; comparisons use the stated controls.\end{minipage}
\end{table*}

\noindent\textbf{Response-format diagnosis.} ScienceQA-IMG changes from 69.61\% for the parent to 54.14\% and 2.28\% after paired 1\% and 5\%. The official extractor rejects lowercase-only option letters. Across all 2,017 questions, the two continuations produce respectively 453 and 1,955 invalid outputs, including 445 and 1,727 standalone lowercase choices. A separate diagnostic that uppercases only these choices gives 67.63\% and 59.20\%; the parent and four instruction controls contain no invalid outputs and remain unchanged. Thus, much of the official-score decrease reflects answer-format drift, with residual choice errors still present. Official scores are retained without correction. This distinction is consistent with the importance of response-format prompting in LLaVA-1.5~\citep{liu2023improvedllava} and motivates testing instruction replay after paired adaptation.

TextVQA also exhibits response changes: 338 of the 856 questions whose score decreases after paired 5\% contain a consensus-supported reference within a longer predicted answer. This descriptive audit uses official answer normalization and requires support from at least two annotators. Such containment is insufficient for a correct response, but the longer answers, together with the ScienceQA format changes, motivate restoring instruction supervision after caption adaptation.

We test whether instruction replay reverses these changes by giving the parent and its paired-1\% descendant the same instruction subset (Appendix~\ref{app:instruction_replay}). All four checkpoints are evaluated under one inference policy. Relative to the paired-only checkpoint, replay eliminates invalid ScienceQA answers and improves official accuracy by 13.98 percentage points without changing the extractor. TextVQA improves by 2.25 points (95\% paired-image interval $[1.47,3.04]$). On MMBench EN, circular accuracy improves by 31.70 points ($[28.87,34.54]$), with unparsed rotations falling from 1,590 to one. Both replayed models score 61.60\%. Instruction replay therefore repairs most format failures and much of the score deterioration. The equally replayed control is essential: the two replayed models' differences on OCRBench, TextVQA, ScienceQA, and MMBench all have intervals containing zero.

\section{Discussion}
\label{sec:discussion}

Inverse-LLaVA shows that pretrained vision and language backbones can be connected through text-to-vision updates within decoder attention, trained on visual instructions alone. The central result is a viable change in mapping direction and fusion location: visual features enter at their encoder dimension, while the language residual path remains intact. The resulting 665K-instruction model approaches two-stage LLaVA-1.5 on several tasks and improves VizWiz over LoRA, with TextVQA and MMBench deficits. Single-stage training is also effective with a conventional projector in our low-data controls. These results support inverse fusion as an alternative interface; single-stage training is feasible with either design.

The supervision interventions reveal a separation between learning visual associations and retaining task-appropriate responses. Celebrity recognition improves with caption pairs, whereas increasing caption exposure alone worsens scene-text QA and response formatting. Instruction replay corrects most format failures and restores much of the benchmark performance. The effects show that the final supervision objective materially influences how the model expresses its knowledge; the remaining TextVQA gap also involves answer choices beyond format alone.

The representation results explain how this alternative connection transforms its inputs. The text maps reduce dimensionality while retaining strong similarity ordering; Q/K contract pairwise distances and V expands them. Early decoder differences give way to similar later geometry, including image-induced activation changes. Thus, distinct modality projections can lead to closely related downstream representations. The measurements concern activation geometry, not equivalence of attention patterns or preservation of all task-relevant information.

The primary results use one training seed; the reported intervals quantify evaluation-sample uncertainty, leaving variation across retraining unmeasured. Component, HD, and 13B studies establish behavior under matched limited-data conditions; their rankings require confirmation at full training scale. Near-floor scores and capacity differences further limit some component comparisons. Training records and inference profiles document the measured costs, while a matched end-to-end training baseline is still needed to quantify wall-clock savings from omitting alignment pretraining.

\section{Conclusion}
\label{sec:conclusion}

Inverse-LLaVA introduces text-to-vision fusion inside decoder attention, connecting frozen pretrained backbones through jointly learned fusion and LoRA parameters. Training on 665K visual instructions approaches two-stage LLaVA-1.5 on several benchmarks without its dedicated 558K-example alignment stage. Component and representation analyses characterize the modality-specific transformations, while cost accounting quantifies their added computation. Supervision interventions identify improved celebrity recognition and reversible answer-format changes at fixed architecture. Together, these results establish a practical alternative to early visual projection and show why its performance must be understood jointly with the supervision objective.

\vspace{-1ex}

\section*{Statements and Declarations}
\vspace{-1ex}
\subsection*{Availability of data and material}
Training and evaluation use the publicly available datasets cited in the manuscript. Code, weights, and reproducibility artifacts are available in the \href{https://github.com/xuhuizhan5/Inverse-LLaVA}{project repository}.
\vspace{-1ex}
\subsection*{Competing interests}
The authors declare that they have no competing interests.
\vspace{-1ex}
\subsection*{Funding}
This research is supported in part by the National Science Foundation (NSF) under grant number IIS2524380. Additionally, this work was supported by a Compute Grant from the Data Science Institute at Vanderbilt University.
\vspace{-1ex}
\begin{samepage}
\subsection*{Authors' contributions}

\noindent Xuhui Zhan: Conceptualization; Methodology; Software; Investigation; Formal analysis;~Writing--original~draft.
\par\nopagebreak[4]

\noindent Tyler Derr: Conceptualization; Methodology; Supervision; Writing--review \& editing;~Funding~acquisition.
\par\nopagebreak[4]

\noindent All authors read and approved the final manuscript.
\end{samepage}
\vspace{-1ex}
\subsection*{Acknowledgements}
The authors would like to thank the Data Science Institute at Vanderbilt University for providing computational resources and support.

\subsection*{Use of AI-assisted tools}
AI-assisted tools were used for code development, analysis scripting, and language editing during manuscript preparation. Experimental claims are based on the model evaluations and analyses reported in this paper. The authors are responsible for the content of the manuscript.

\clearpage
\begin{appendices}
\renewcommand{\theHfigure}{appendix.\thefigure}
\renewcommand{\theHtable}{appendix.\thetable}
\renewcommand{\theHequation}{appendix.\theequation}
\section{Additional Benchmark Results}
\label{app:local_inventory}

Table~\ref{tab:canonical_panel} includes the local benchmark comparisons and supplementary tasks under their stated metrics. Each Inverse-LLaVA entry uses the final checkpoint; MME cognition is reported separately from perception.

\begin{table}[!ht]
\centering
\small
\caption{Additional comparisons with official LLaVA-1.5 checkpoints on complete splits. Scores are percentages except MME; MMStar uses macro-L2 accuracy. All entries use BF16/SDPA on Thor. $N$ is the number of questions. Model and training differences follow Section~\ref{sec:setup}.}
\label{tab:canonical_panel}
\begin{tabular}{lr>{\columncolor{methodshade}}rrr}
\toprule
Benchmark & $N$ & Inverse-LLaVA & LoRA & FFT \\
\midrule
OCRBench & 1,000 & 31.00 & \textbf{32.40} & \underline{31.70} \\
MME cognition & 260 & \underline{279.29} & 258.21 & \textbf{344.64} \\
MME perception & 2,114 & 1453.82 & \underline{1484.58} & \textbf{1507.28} \\
TextVQA val & 5,000 & 56.96 & \textbf{58.47} & \underline{58.21} \\
ScienceQA-IMG & 2,017 & \textbf{69.61} & 68.82 & \underline{69.11} \\
AI2D & 3,088 & 54.53 & \textbf{55.31} & \underline{55.28} \\
MMStar & 1,500 & \underline{35.17} & \textbf{35.24} & 33.61 \\
GQA test-dev & 12,578 & \underline{62.28} & \textbf{62.63} & 61.89 \\
\bottomrule
\end{tabular}
\par\smallskip\begin{minipage}{0.98\linewidth}\footnotesize Re-evaluated row best/second: bold/underline. Gray: Inverse-LLaVA.\end{minipage}
\end{table}
\FloatBarrier

\subsection{Paired differences and response diagnostics}
\label{app:benchmark_differences}

On all 107,394 VQAv2 test-dev questions, official server scores are 78.45\% for Inverse-LLaVA, 79.13\% for LoRA, and 78.55\% for FFT. The point differences are $-0.68$ and $-0.10$ percentage points, respectively. The server returns aggregate scores without per-question labels, so we do not attach itemwise bootstrap intervals to these results.

On all 8,000 VizWiz test questions, Inverse-LLaVA scores 50.96\%, compared with 48.56\% for LoRA and 50.64\% for FFT. Paired bootstrap intervals over 7,928 decoded-image groups give differences of $+2.40$ percentage points ($[1.58,3.21]$) and $+0.32$ points ($[-0.50,1.15]$), respectively. An exploratory analysis using the released answerability labels locates the LoRA-relative gain in the 2,147 unanswerable questions: Inverse scores 70.04\%, versus LoRA 56.91\% and FFT 76.41\%. On the 5,853 answerable questions, scores are 43.96\%, 45.50\%, and 41.19\%. The aggregate ranking therefore combines different answering and abstention behavior.

On MM-Vet, Inverse-LLaVA scores 28.67\%, compared with 31.24\% for LoRA and 29.68\% for FFT under the same official hosted judge. The paired differences are $-2.57$ percentage points ($[-6.03,0.96]$) and $-1.01$ points ($[-4.37,2.51]$). Both intervals include zero and condition on one judging run, excluding judge-run variability.

On MMBench EN dev, circular accuracy is 62.63\% for Inverse-LLaVA, 67.10\% for LoRA, and 65.12\% for FFT. Paired differences over the 1,164 question groups are $-4.47$ percentage points ($[-6.27,-2.66]$) and $-2.49$ points ($[-4.38,-0.60]$). All 4,329 rotated answers per model are parsed by the declared extractor, ruling out unparsed responses as an explanation of these differences. Accuracy on the base rotation alone is 72.16\%, 74.91\%, and 74.48\%, respectively. The gap is present before circular scoring and increases when all rotations must be correct.

On MMBench CN dev, the corresponding scores are 54.04\%, 58.93\%, and 58.33\%. Paired differences are $-4.90$ percentage points ($[-6.87,-2.92]$) and $-4.30$ points ($[-6.53,-2.15]$). Inverse has 43 unparsed rotations, versus one for LoRA and three for FFT. Even granting every unparsed Inverse answer full credit could recover only seven circular groups, or 0.60 points. Thus extraction failures alone cannot close the observed CN gap; the reported scores remain unchanged.

On the complete GQA test-dev-balanced split, Inverse-LLaVA scores 62.28\%, compared with 62.63\% for LoRA and 61.89\% for FFT. Resampling the 398 image groups while retaining question-weighted accuracy gives paired differences of $-0.35$ percentage points ($[-0.98,0.27]$) and $+0.39$ points ($[-0.23,0.98]$), respectively. Both intervals include zero.

\subsection{Published and Re-evaluated Contemporary References}
\label{app:contemporary_reference}

The published scores and common-task comparison in Section~\ref{sec:canonical_evaluation} span different visual encoders, language backbones, and training recipes. Table~\ref{tab:model_context} details these differences. Appendix~\ref{app:matched_interface} compares interfaces at matched instruction exposure and CLIP features.

We evaluate official Qwen3-VL-8B-Instruct~\citep{bai2025qwen3vl}, revision \texttt{0c351dd01ed8}, on complete MMBench EN dev. Questions, options, images, and circular groups match the primary comparison. Inference uses its native image processor and chat template, BF16, SDPA, batch size one, greedy decoding, and the same 1,024-token limit.

Qwen3-VL reaches 84.97\% circular accuracy. Paired gaps relative to Inverse-LLaVA, LoRA, and FFT are $+22.34$ points ($[19.59,25.00]$), $+17.87$ ($[15.29,20.53]$), and $+19.85$ ($[17.18,22.51]$), respectively, using the question-group bootstrap in Appendix~\ref{app:reproduction_protocols}.

The pinned upstream extractor agrees on all 17,316 responses. Qwen3-VL has 11 unparsed rotations; correcting all of them could add at most 0.26 points. Its stronger system-level performance includes the training and architectural differences in Table~\ref{tab:model_context}.
\begin{table*}[!ht]
\centering
\small
\caption{Interfaces and training regimes of the compared model families~\citep{liu2023improvedllava,liu2024llavanext,karamcheti2024prismatic,tong2024cambrian,wang2024qwen2vlenhancingvisionlanguagemodels,qwen2025qwen25vl,bai2025qwen3vl}. Training counts exclude inherited backbone pretraining. In the training column, K/M count examples and B/T count tokens. Arrows show training order, not equivalent compute. Gray: Inverse-LLaVA.}
\label{tab:model_context}
\setlength{\tabcolsep}{4pt}
\renewcommand{\arraystretch}{1.15}
\begin{tabular}{@{}>{\raggedright\arraybackslash}p{2.1cm}>{\raggedright\arraybackslash}p{3.0cm}>{\raggedright\arraybackslash}p{2.5cm}>{\raggedright\arraybackslash}p{4.5cm}@{}}
\toprule
Model & Language and vision & Interface & VLM supervision and adaptation \\
\midrule
\rowcolor{methodshade}
Inverse-LLaVA & Vicuna-7B; CLIP-L/14, 336 px & Text-to-vision Q/K/V & 665K instructions; joint fusion + LoRA \\
\addlinespace[3pt]
LLaVA-1.5 & Vicuna-7B; CLIP-L/14, 336 px & Early visual MLP & 558K alignment $\to$ 665K instructions; LoRA or FFT \\
\addlinespace[3pt]
Prismatic 7B (one-stage) & Vicuna-7B; CLIP-L/14, 336 px & Early GELU MLP & LLaVA instructions; joint projector + full LLM adaptation \\
\addlinespace[3pt]
LLaVA-1.6 7B & Vicuna-7B; CLIP-L/14, 336-px AnyRes crops & Early visual MLP & 558K alignment $\to$ 760K instructions; document/chart and GPT-4V data; FFT \\
\addlinespace[3pt]
Cambrian-1-8B & Llama-3-8B-Instruct; four encoders, 336--1024 px & Spatial Vision Aggregator & 2.5M connector data $\to$ 7M instructions; 576 visual tokens \\
\addlinespace[3pt]
Qwen2-VL-7B & Qwen2; trainable 675M ViT; dynamic resolution & Spatial MLP merger; M-RoPE & 1.4T pretraining tokens $\to$ instruction tuning; OCR, documents, video \\
\addlinespace[3pt]
Qwen2.5-VL-7B & Qwen2.5; dynamic ViT; window attention & Spatial MLP merger; time-aware M-RoPE & 4.1T tokens in three pretraining stages $\to$ supervised tuning; OCR, grounding, documents, video \\
\addlinespace[3pt]
Qwen3-VL family & Qwen3; dynamic-resolution SigLIP-2 & MLP merger; DeepStack & Pretraining phases: 67B, ${\sim}1$T, ${\sim}1$T, 100B tokens $\to$ post-training \\
\bottomrule
\end{tabular}
\end{table*}
\FloatBarrier

\begin{figure}[H]
\centering
\includegraphics[width=\textwidth]{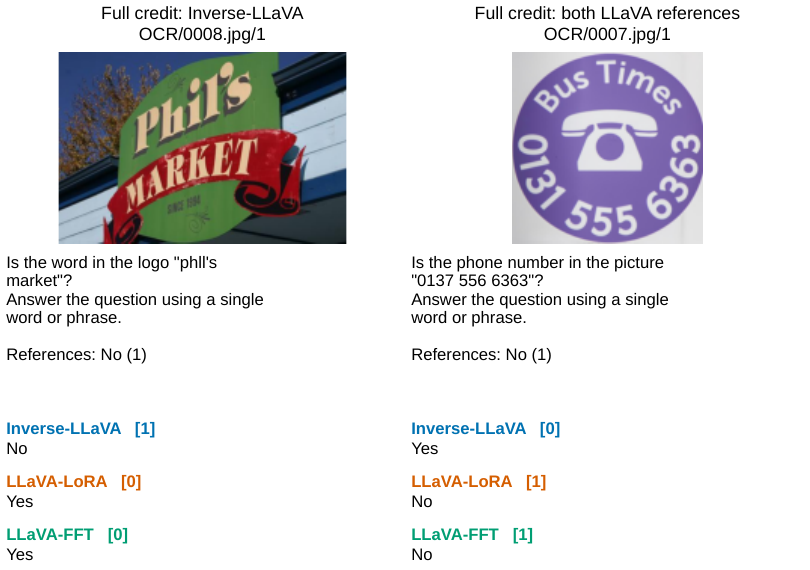}
\caption{Matched MME OCR examples~\citep{fu2023mme}. The negative references concern a misspelled name and an incorrect phone number. Brackets give official per-question scores. Checking the companion positive questions, Inverse-LLaVA answers both logo questions correctly, and FFT answers both phone-number questions correctly. LoRA answers ``No'' to both phone-number questions, so its displayed rejection does not imply pair-level correctness. Prompts and answers are unedited apart from omission of the fixed chat wrapper. These selected cases illustrate behavior without establishing its cause.}
\label{fig:mme_ocr_cases}
\end{figure}

\begin{figure}[H]
\centering
\includegraphics[width=\textwidth]{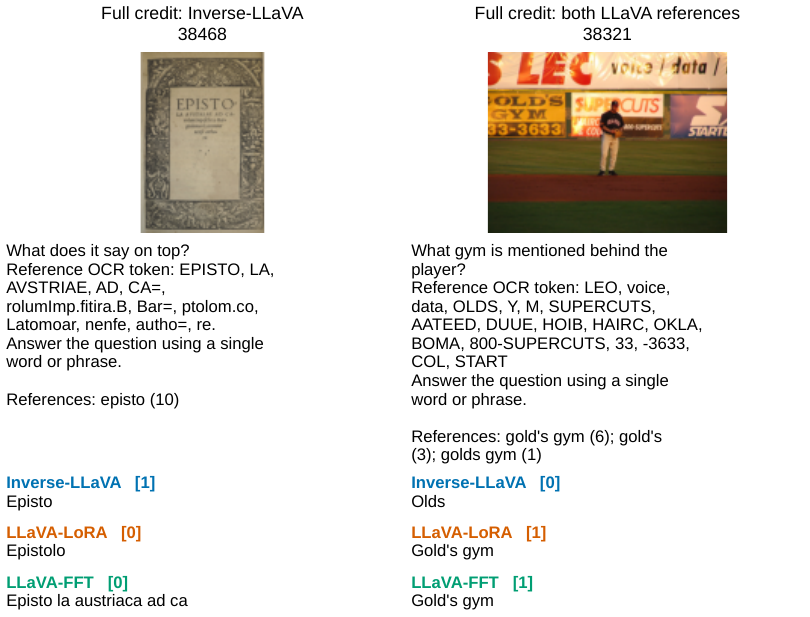}
\caption{Matched TextVQA examples~\citep{singh2019textvqa}, including the OCR hints supplied to every model. Inverse-LLaVA answers the word-reading question correctly; both LLaVA references answer the gym-name question correctly. Brackets give official consensus item scores, and reference parentheses count annotators. Answers receive the standard normalization and scoring, without manual correction. Individual examples complement the full-validation scores and do not establish an OCR mechanism.}
\label{fig:textvqa_cases}
\end{figure}

\begin{samepage}
\section{Qualitative Examples}
\label{app:benchmark_cases}

\subsection{Visual questions and scientific reasoning}

Figures~\ref{fig:vizwiz_cases} and~\ref{fig:scienceqa_cases} extend the comparison to VizWiz and ScienceQA-IMG. For each benchmark, the complete casebook retains three first-hash-ranked examples from each of six full-credit outcome groups (seed 2026). VizWiz is additionally stratified by released answerability labels, giving 36 cases; ScienceQA contributes 18. The figure uses the first case where only Inverse-LLaVA receives full credit and the first where both references receive full credit. VizWiz's displayed cases are from its answerable stratum; unanswerable examples remain in the casebook. Partial-credit responses retain their numerical scores. The deterministic selections illustrate contrasting responses without estimating their prevalence.

\begin{figure}[!ht]
\centering
\includegraphics[width=\textwidth]{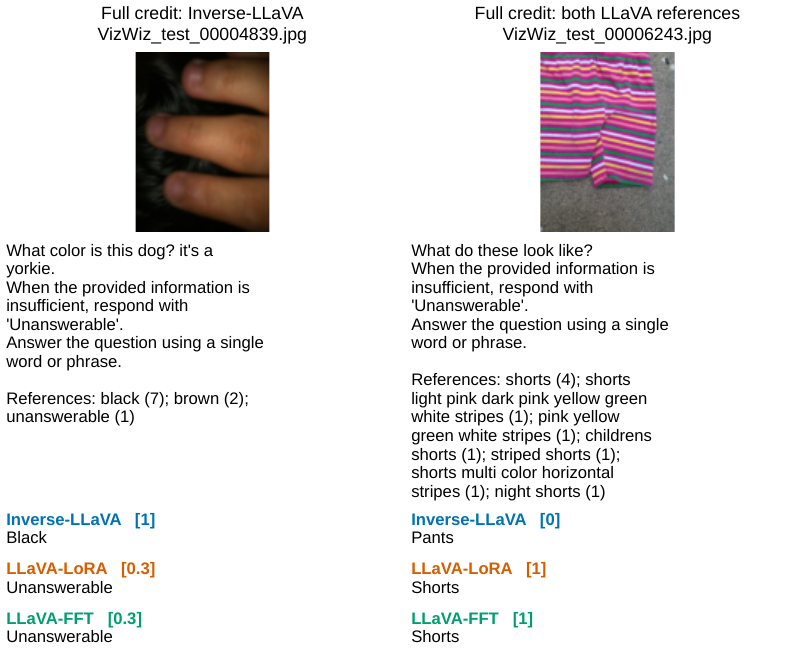}
\caption{Matched answerable VizWiz examples from the complete primary test evaluation. The same image, question, and scoring rules apply to all three models. Brackets give official consensus item scores; reference parentheses count annotators. Answers are unedited. These selected examples complement the answerability-stratified aggregate analysis in Appendix~\ref{app:benchmark_differences}.}
\label{fig:vizwiz_cases}
\end{figure}

\begin{figure}[!ht]
\centering
\includegraphics[width=\textwidth]{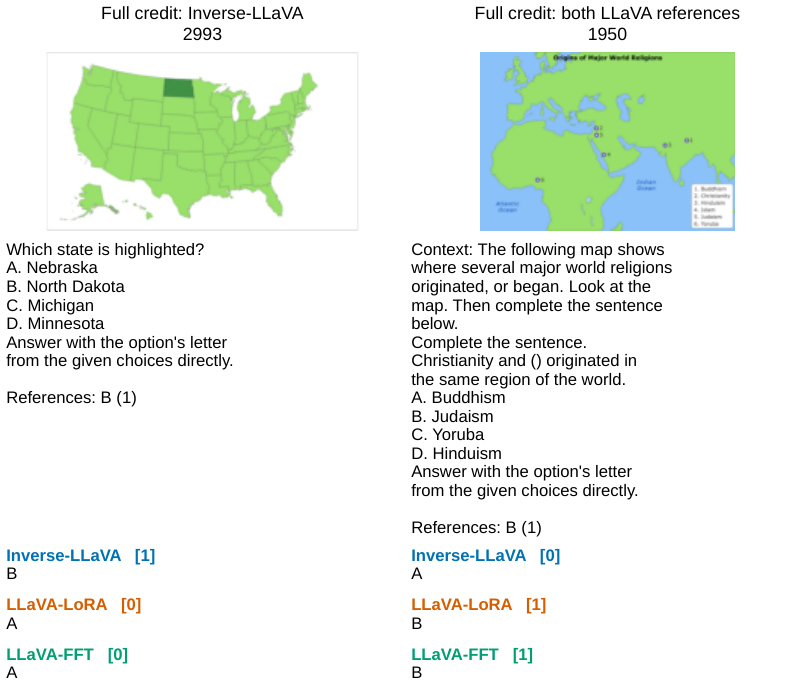}
\caption{Matched ScienceQA-IMG examples from the complete primary test evaluation, including the question, choices, and option-letter instruction. The final Inverse-LLaVA model succeeds on the left example and both official references succeed on the right. All displayed answers are valid choices; brackets show official item scores. These cases illustrate answer differences without revealing the models' internal reasoning.}
\label{fig:scienceqa_cases}
\end{figure}

\FloatBarrier
\subsection{Text recognition and answering}

Figures~\ref{fig:mme_ocr_cases} and~\ref{fig:textvqa_cases} use predictions from the complete primary evaluation splits. For each task, we rank examples by a fixed seed-2026 hash within three-model full-credit patterns and display the first example where only Inverse-LLaVA receives full credit and the first where both LLaVA references do. The accompanying casebooks retain up to three examples from all six outcome patterns: 16 MME OCR cases and 18 TextVQA cases. TextVQA's continuous item scores and all ten annotations are preserved. The selection illustrates contrasting outputs; it does not estimate the frequency of errors.
\par
\end{samepage}

\noindent\textbf{Supplementary image interventions.}\par\nopagebreak[4]

We also freeze an 18-example OCRBench panel, sampling three examples from each of six three-model outcome groups with seed 2026. Replacing each image with a same-size uniform image or a deranged image assignment makes all nine originally correct Inverse-LLaVA answers incorrect. Under the same interventions, LoRA changes from ten correct answers to one (all ten original successes are lost, with one newly correct response), and FFT changes from eight to zero. These outcome-balanced examples document image dependence; they do not estimate benchmark accuracy or isolate an error mechanism. The full original, blank-image, and shuffled-image outputs remain in the accompanying casebook.

\section{Representation Geometry}
\label{app:canonical_geometry}

\begin{figure*}[!t]
\centering
\includegraphics[width=\textwidth]{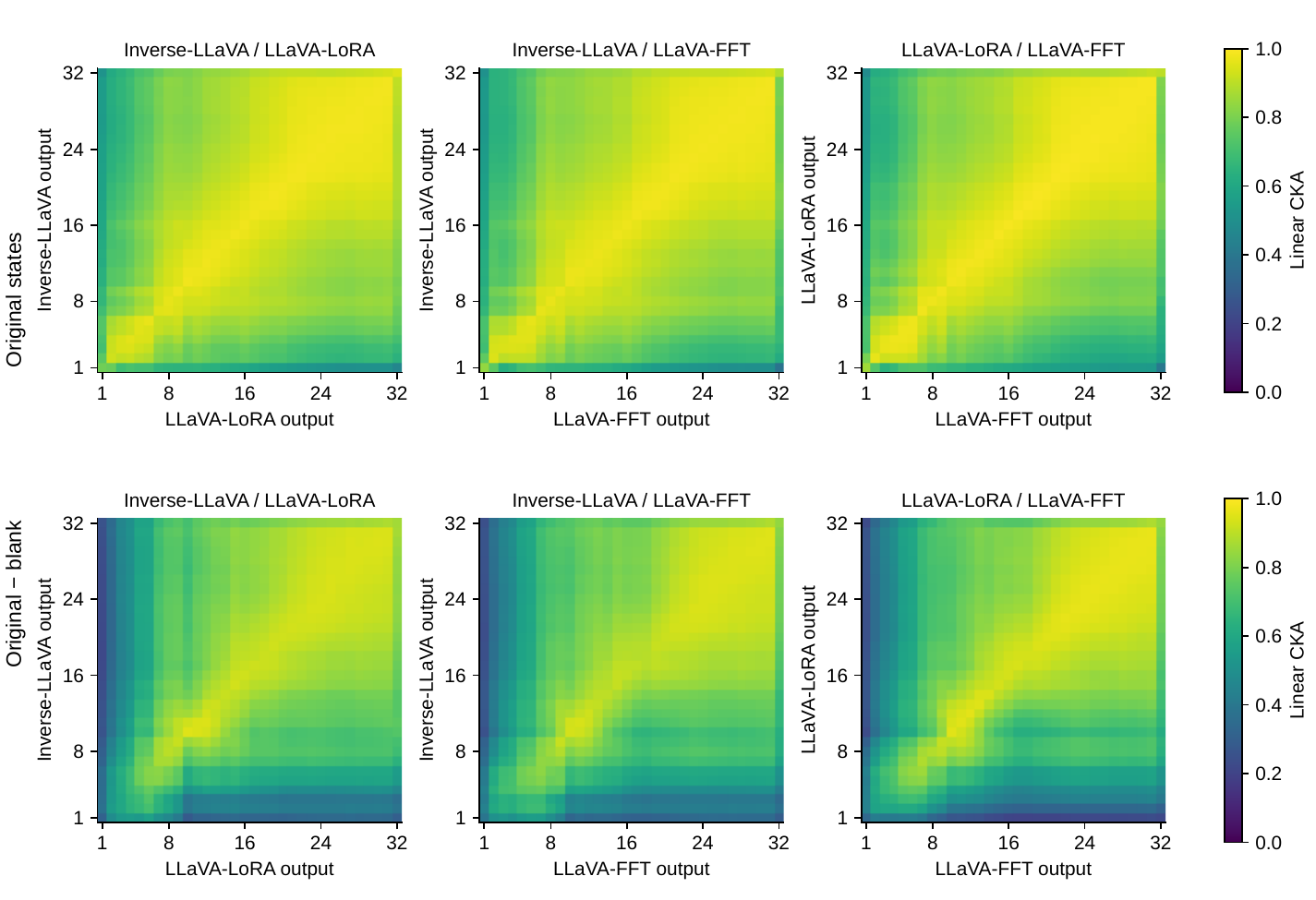}
\caption{Cross-layer linear CKA at the final valid prompt-token position: original states (top) and within-model original-minus-blank activation differences (bottom). Matrix rows refer to the first named model and columns to the second. All panels use the same 100 image/question pairs and a common color scale. The intervention complements original-state similarity, which also reflects shared prompts and pretrained structure.}
\label{fig:canonical_geometry}
\end{figure*}

All representation figures and metrics use one ordered panel of 100 OCRBench examples. We rank sample IDs by a seed-2026 hash and retain the first unused question and decoded RGB image, without consulting predictions. Questions are deduplicated after Unicode, whitespace, and case normalization. The resulting panel contains 41 document-oriented VQA, 30 scene-text VQA, 26 key-information extraction examples, and one each from three recognition categories. This question-diverse diagnostic differs from the benchmark's category distribution. All 100 prompts and images, and all 100 mapped text vectors in each branch, are distinct.

Figure~\ref{fig:canonical_geometry} compares final valid prompt-token states across models. Output indices 1--32 identify decoder block outputs; index 0 is excluded because the final prompt-token embedding is constant across samples. Captures use batch size one and BF16/SDPA on Thor. All nine declared pair/layer comparisons at outputs 1, 16, and 32 exceed a row-permutation null with Holm-adjusted $p<0.005$ (2,000 permutations). The null tests sample correspondence, including question content, rather than visual grounding. Bootstrap intervals condition on these fitted checkpoints and exclude training-seed uncertainty. Repeated rows in a bootstrap resample can inflate high-dimensional CKA, motivating the estimator check below.

To examine image-dependent changes, we repeat each capture with a same-size blank image while preserving its question, then compute the per-example difference between original and blank activations. CKA between these differences is 0.316 for Inverse--LoRA at output 1, 0.926 at output 16, and 0.862 at the final output. Against FFT, the corresponding values are 0.388, 0.908, and 0.836. This control describes responses to an image intervention, including nonlinear interactions with the question. It assumes no additive decomposition of hidden states into vision and language.

A finite-sample sensitivity check normalizes unbiased HSIC estimates using the reference formulation of \citet{kornblith2019similarity}. The resulting CKA ratio can still be biased. All 18 original-state and image-change comparisons reproduce the authors' implementation within $3\times10^{-15}$. At the final output, Inverse--LoRA and Inverse--FFT original-state estimates change from 0.950/0.885 to 0.942/0.883; image-change estimates change from 0.862/0.836 to 0.776/0.739. Early image-change similarities become 0.272/0.357 at output 1, while output-16 values remain 0.903/0.882. Thus the early-versus-middle pattern persists, with estimator-sensitive magnitudes. The ordinary-CKA confidence intervals do not apply here.

\subsection{Before and after the modality projections}
\label{app:projection_geometry}

We reuse the same 100 paired captures to examine transformations within the trained models. For Inverse-LLaVA, the pre-map text direction is the pooled decoder input, whose image slots are zero; the post-map representation includes per-token RMSNorm and the branch-specific mapper. For LLaVA, we compare pooled selected CLIP features with their projected outputs. Correspondence is by example, so relational metrics can compare different feature widths without fitting a cross-model alignment. All metrics operate on L2-normalized high-dimensional vectors.

\begin{table}[!ht]
\centering\small
\caption{Within-model geometry on 100 fixed examples. RSA is the Spearman correlation of pairwise cosine similarities; kNN-10 is the overlap of ten-nearest-neighbor sets; CKA is linear centered kernel alignment. The distance ratio is the median pairwise cosine distance after transformation divided by that before transformation. Gray identifies the inverse text branches.}
\label{tab:projection_geometry}
\begin{tabular}{lrrrrr}
\toprule
Transformation & Dimension & RSA & CKA & kNN-10 & Ratio \\
\midrule
\rowcolor{methodshade}
Inverse text, Q branch & $4096\to1024$ & .803 & .790 & .560 & .040 \\
\rowcolor{methodshade}
Inverse text, K branch & $4096\to1024$ & .810 & .748 & .581 & .013 \\
\rowcolor{methodshade}
Inverse text, V branch & $4096\to1024$ & .831 & .847 & .619 & 4.163 \\
LoRA visual projector & $1024\to4096$ & .902 & .895 & .746 & 1.261 \\
FFT visual projector & $1024\to4096$ & .839 & .789 & .686 & .529 \\
\bottomrule
\end{tabular}
\end{table}

\begin{figure}[!ht]
\centering
\includegraphics[width=\textwidth]{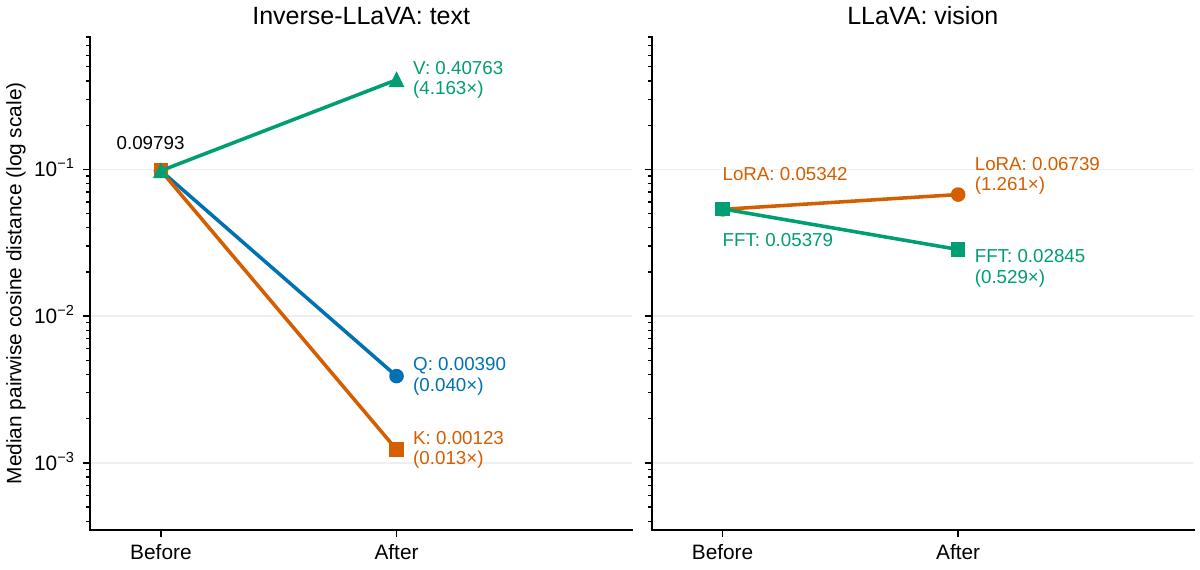}
\caption{Pairwise distances before and after modality projection. Points show median cosine distance across the same 4,950 example pairs; parentheses give after/before ratios. Both panels use the same logarithmic scale. Q/K text branches contract and V expands, despite similar preservation of similarity ordering (Table~\ref{tab:projection_geometry}). The inverse transformations include normalization.}
\label{fig:projection_geometry}
\end{figure}

\begin{figure}[!ht]
\centering
\includegraphics[width=\textwidth]{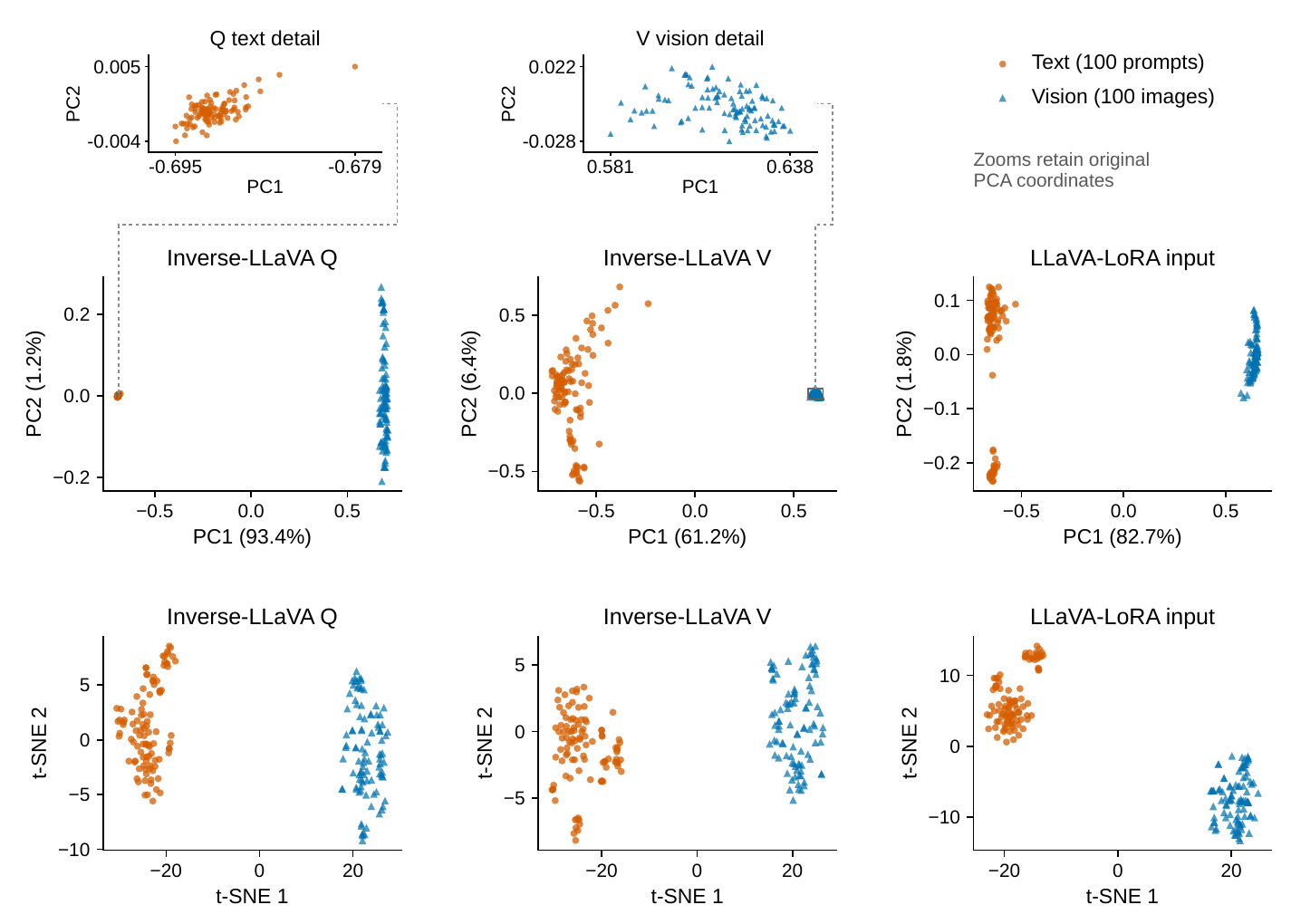}
\caption{Modality distributions in PCA (middle) and t-SNE (bottom; perplexity 30, seed 2026)~\citep{vandermaaten2008visualizing}. Each fit combines the same 100 distinct text and 100 distinct visual vectors after row normalization, without separate modality centering. Dashed guides link the boxed Q text and V vision clusters to zooms in the same PCA coordinates. Axis percentages report explained variance. Inverse-LLaVA uses mapped text and normalized vision; LoRA uses input text and projected vision. Panels have independent coordinates, and t-SNE spacing is not comparable to PCA spacing. Table~\ref{tab:projection_geometry} quantifies the geometry in the original feature spaces.}
\label{fig:joint_modalities}
\end{figure}

The inverse text input has median cosine distance 0.09793, compared with 0.00390/0.00123/0.40763 after Q/K/V transformation. The RSA values describe substantial preservation of similarity ordering despite changes in pairwise distances. LoRA's visual median changes from 0.05342 to 0.06739, while FFT's changes from 0.05379 to 0.02845. In the joint Q view, PC1 explains 93.4\% of variance and primarily separates the modality means. The Q text vectors remain tightly clustered even with distinct questions; the inset reveals their residual variation. These pooled, pre-attention features include the shared chat wrapper. The plots characterize the trained interfaces; a training trajectory would require captures across training in common coordinates. Additional K-branch views and t-SNE sensitivity checks accompany the analysis.

In the inverse visual branch, offline reconstruction through the Q/K/V output maps gives native-to-output RSA values of 0.750/0.686/0.939 and kNN-10 overlaps of 0.597/0.565/0.807. These estimates describe the pooled visual contribution before RoPE and attention, reconstructed in FP32. The combined pattern is branch-specific re-encoding with substantial relational continuity. Pooling, question content, chat-wrapper content, CLIP-layer selection, and normalization affect these measurements; semantic-information preservation requires a task-linked test.

\FloatBarrier
\section{Interface and Model Size}
\label{app:matched_interface}

\subsection{Interface and CLIP-Layer Selection}
\label{app:feature_selection}

We train Inverse-LLaVA and a two-layer early visual projector with either final or penultimate CLIP patch features. All four conditions use the same 33,265 instruction rows, 260-update cosine schedule, global batch 128, and shared LoRA initialization, excluding decoder layer 0 from LoRA. The inverse conditions also share their fusion initialization; the projector conditions share a freshly initialized projector, without alignment pretraining. All backbones remain frozen. Trainable counts are 347,573,251 for inverse fusion and 330,801,152 for the early projector. Thus, supervision and adaptation placement are controlled while interface capacity differs. These projector controls are distinct from the official two-stage LLaVA checkpoints.

\begin{table}[!ht]
\centering
\small
\setlength{\tabcolsep}{3.5pt}
\caption{Matched single-stage interface/CLIP-layer study: 260 updates on the same 5\% instruction subset, global batch 128, and one seed. Both interfaces train LoRA jointly; the early projector has no alignment pretraining. OCRBench, TextVQA, and ScienceQA-IMG use fixed 256-question subsets; MME categories use complete question pairs. QA values are percentages; MME values are benchmark scores. Final update 260 is primary.}
\label{tab:matched_interface}
\begin{tabular}{llrrrrr}
\toprule
Model & CLIP features & \shortstack{OCRBench} & \shortstack{TextVQA} & \shortstack{ScienceQA\\IMG} & \shortstack{MME\\cognition} & \shortstack{MME\\OCR} \\
\midrule
\rowcolor{methodshade}
Inverse-LLaVA & Final & 1.95 & 38.12 & 58.20 & 296.07 & 95.00 \\
Early projector & Final & 2.73 & 38.83 & 58.59 & 291.79 & 87.50 \\
\midrule
\rowcolor{methodshade}
Inverse-LLaVA & Penultimate & 3.52 & 37.73 & 63.28 & 262.14 & 102.50 \\
Early projector & Penultimate & 2.34 & 38.79 & 61.33 & 283.21 & 65.00 \\
\bottomrule
\end{tabular}
\par\smallskip\begin{minipage}{0.98\linewidth}\footnotesize Each block uses the same CLIP features. Gray: Inverse-LLaVA.\end{minipage}
\end{table}

Penultimate-minus-final ScienceQA-IMG changes are $+5.08$ percentage points for Inverse-LLaVA (95\% paired interval $[0.77,9.65]$) and $+2.73$ for the projector ($[0.76,5.08]$). All 1,024 ScienceQA responses are valid option labels. The interaction, defined as the inverse feature effect minus the projector feature effect, is $+2.34$ points ($[-2.36,7.17]$). Its interval includes zero, so the observed benefit cannot be attributed specifically to inverse fusion. Inverse-LLaVA's TextVQA change is $-0.39$ points ($[-3.63,2.94]$). This does not meet the prespecified criterion for a larger follow-up: a two-point TextVQA gain with an interval above zero and no decline exceeding one point on OCRBench or ScienceQA. The primary model keeps the final CLIP layer.

All ten inverse-minus-projector intervals and all five interaction intervals include zero. OCRBench is near floor, and MME OCR contains 20 image pairs. All comparisons in this appendix use 10,000 paired resamples with seed 2026: QA resamples decoded-image groups; MME resamples complete image pairs within categories. Intervals are conditional on these single-seed checkpoints, unadjusted for multiplicity, and based on reused development subsets. They do not establish equivalence or predict a full-data ranking. Complete contrasts and response audits accompany each study.

\FloatBarrier
\subsection{Visual Feature Width}
\label{app:matched_width}

The width study compares Inverse-LLaVA's 1,024-dimensional final CLIP features with Inverse-LLaVA-HD's 2,048-dimensional penultimate-plus-final features. Both retain 336-pixel input resolution, 576 patches, and the 7B language backbone. They use the same 33,265 selected instruction examples, 33,280 sampled examples including padding, global batch 128, and 260-update schedule. Their trainable counts are 347,573,251 and 385,325,059, respectively. We evaluate updates 100, 200, and 260 on common fixed inputs, using update 260 for the final comparison. HD changes feature content as well as dimension and capacity, so this comparison tests the higher-dimensional configuration as a whole.

\begin{table}[!ht]
\centering
\small
\setlength{\tabcolsep}{3.5pt}
\caption{Matched 5\% instruction-subset width study, one seed and a shared 260-update schedule. HD changes visual feature content and width. OCRBench, TextVQA, and ScienceQA-IMG use fixed 256-question subsets; MME categories use complete question pairs. QA values are percentages; MME values are benchmark scores. Final update 260 is primary.}
\label{tab:matched_width}
\begin{tabular}{llrrrrr}
\toprule
Model & Update & \shortstack{OCRBench} & \shortstack{TextVQA} & \shortstack{ScienceQA\\IMG} & \shortstack{MME\\cognition} & \shortstack{MME\\OCR} \\
\midrule
Inverse-LLaVA & 100 & 1.56 & 36.45 & 55.86 & 252.50 & 65.00 \\
Inverse-LLaVA-HD & 100 & 1.56 & 38.20 & 56.64 & 222.86 & 50.00 \\
\midrule
Inverse-LLaVA & 200 & 2.34 & 39.06 & 59.77 & 312.14 & 100.00 \\
Inverse-LLaVA-HD & 200 & 2.34 & 38.67 & 60.94 & 305.00 & 87.50 \\
\midrule
Inverse-LLaVA & 260 & 1.95 & 38.12 & 58.20 & 296.07 & 95.00 \\
Inverse-LLaVA-HD & 260 & 2.34 & 39.06 & 60.55 & 283.57 & 102.50 \\
\bottomrule
\end{tabular}
\par\smallskip\begin{minipage}{0.98\linewidth}\footnotesize Each block compares models at the same optimizer update.\end{minipage}
\end{table}

At update 260, HD-minus-standard differences are $+0.94$ TextVQA percentage points (95\% paired interval $[-1.84,3.82]$), $+2.34$ ScienceQA-IMG points ($[-1.94,6.61]$), $+0.39$ OCRBench points ($[0.00,1.19]$), $-12.50$ MME cognition points ($[-55.36,31.07]$), and $+7.50$ MME OCR points ($[-15.00,30.00]$). All include zero. At update 100, HD cognition is lower by 29.64 points ($[-59.64,-3.93]$); both models improve by update 200 and have lower cognition point scores at 260. Thus the trajectories are not uniformly monotonic, and the final comparison establishes neither superiority nor equivalence. OCRBench remains near floor, and MME OCR contains only 20 image pairs.

\begin{figure*}[!ht]
\centering
\includegraphics[width=\textwidth]{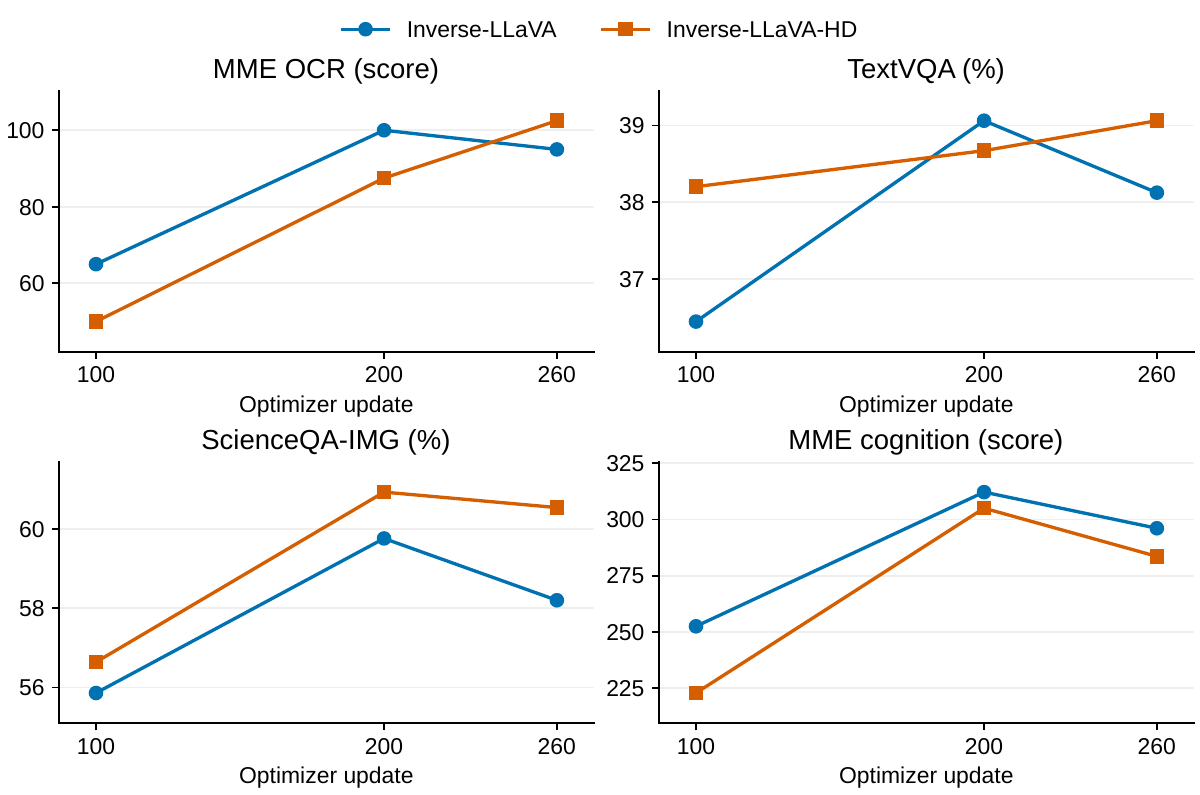}
\caption{Matched Inverse-LLaVA and Inverse-LLaVA-HD checkpoint evaluations on the same 5\% instruction subset and schedule. Lines connect observed scores and are not fitted scaling laws. All three saved milestones are retained; update 260 is primary. Table~\ref{tab:matched_width} also reports OCRBench. Paired comparison intervals are given in the text and complete result record.}
\label{fig:matched_width}
\end{figure*}

The standard model's 1,068 update-260 answers repeat the earlier interface evaluation exactly. All three milestones and their paired contrasts are retained. The trajectories couple exposure, optimization, and schedule, following the uncertainty protocol in Appendix~\ref{app:feature_selection}.

\FloatBarrier
\subsection{Language Backbone Size}
\label{app:matched_language}

We compare Vicuna-7B and Vicuna-13B under the same 33,265-row instruction subset, sample order, 260-update schedule, and final-layer CLIP features. Both jointly train fusion and rank-128 LoRA from untouched initializations, with frozen backbones. The 13B decoder has width 5,120 and 40 layers, versus 4,096 and 32 for 7B; trainable parameters increase from 347,573,251 to 535,366,659. Pretrained weights also differ, so this tests the larger configuration rather than isolating parameter count.

A full-length microbatch-32 test exceeded memory for 13B. Both runs therefore use microbatch 16, two GPUs, and four accumulation steps, retaining global batch 128. We train a fresh 7B control with the same partition because averaging per-microbatch token losses need not equal a globally token-weighted loss. Selected examples, supervised-target counts, and learning rates match at every update. Each run processes 33,280 sampled examples and 27,958,450 expanded tokens. Training cost is 2.15 GPU-hours for 7B and 3.36 for 13B, including loading and checkpointing; peak allocated memory is 38.02 and 57.02 GiB per GPU. Costs omit allocation and validation after training.

\begin{table}[!ht]
\centering
\small
\setlength{\tabcolsep}{3.5pt}
\caption{Matched 7B/13B language-backbone study on the same 5\% instruction subset, one seed and a shared 260-update schedule. Both use final-layer CLIP features, microbatch 16, and global batch 128. Backbone size and pretrained weights differ. OCRBench, TextVQA, and ScienceQA-IMG use fixed 256-question subsets; MME categories use complete question pairs. QA values are percentages; MME values are benchmark scores. Final update 260 is primary.}
\label{tab:matched_language}
\begin{tabular}{llrrrrr}
\toprule
Model & Update & \shortstack{OCRBench} & \shortstack{TextVQA} & \shortstack{ScienceQA\\IMG} & \shortstack{MME\\cognition} & \shortstack{MME\\OCR} \\
\midrule
Inverse-LLaVA 7B & 100 & 1.56 & 38.32 & 57.03 & 288.93 & 105.00 \\
Inverse-LLaVA 13B & 100 & 3.52 & 41.76 & 61.72 & 240.36 & 125.00 \\
\midrule
Inverse-LLaVA 7B & 200 & 2.34 & 38.55 & 58.98 & 283.57 & 122.50 \\
Inverse-LLaVA 13B & 200 & 3.12 & 43.87 & 62.11 & 267.50 & 107.50 \\
\midrule
Inverse-LLaVA 7B & 260 & 1.95 & 39.34 & 57.81 & 280.71 & 77.50 \\
Inverse-LLaVA 13B & 260 & 3.52 & 42.27 & 61.72 & 234.29 & 57.50 \\
\bottomrule
\end{tabular}
\par\smallskip\begin{minipage}{0.98\linewidth}\footnotesize Each block compares models at the same optimizer update.\end{minipage}
\end{table}

\begin{figure*}[!ht]
\centering
\includegraphics[width=\textwidth]{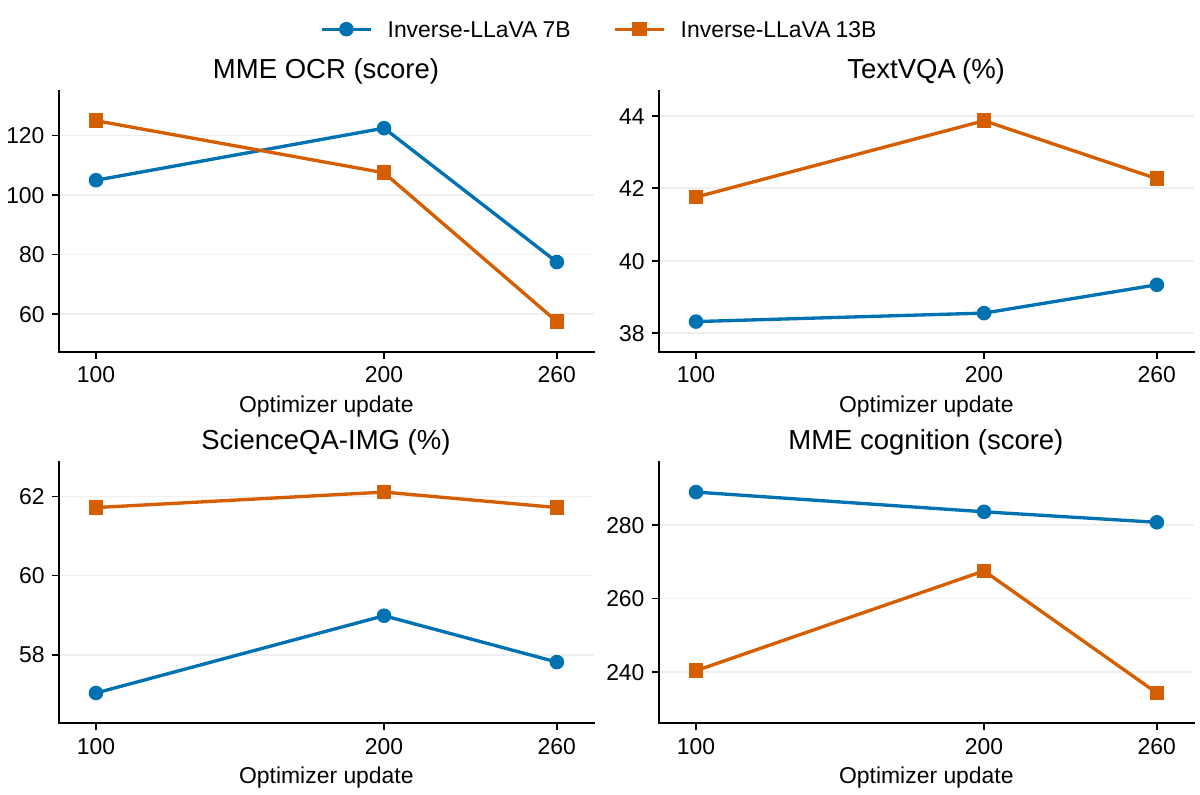}
\caption{Matched language-backbone trajectories at fixed instruction exposure. Both sizes use the same batch partition and training schedule. Lines connect observed checkpoints, with update 260 retained as primary; they are not fitted scaling laws. Table~\ref{tab:matched_language} includes the supplementary OCRBench endpoint. Paired intervals condition on these fitted checkpoints.}
\label{fig:matched_language}
\end{figure*}

At the primary update 260, 13B-minus-7B differences are $+2.93$ TextVQA percentage points (95\% paired interval $[-1.44,7.39]$), $+3.91$ ScienceQA-IMG points ($[-1.95,9.88]$), $+1.56$ OCRBench points ($[0.00,3.53]$), $-46.43$ MME cognition points ($[-82.50,-12.49]$), and $-20.00$ MME OCR points ($[-52.50,10.00]$). TextVQA and ScienceQA point scores are higher for 13B at all three saved milestones, while MME trajectories differ (Figure~\ref{fig:matched_language}). We retain the final checkpoint rather than selecting update 200 after seeing its scores.

The cognition difference is concentrated in Chinese text translation: 7B scores 100.00 and 13B 55.00, accounting for 45.00 of the 46.43-point gap. Their answers contain 30/40 and 38/40 ``No'' responses, respectively, with no invalid labels. Both answer ``No'' to every code and calculation question. Thus, this low-data aggregate also reflects category-specific answer patterns. Lower later training loss for 13B is insufficient to predict its benchmark ranking; the matched raw loss histories and all response audits accompany the result record.

ScienceQA labels are valid at every milestone. All thirty endpoints and thirty-five paired contrasts follow the uncertainty protocol in Appendix~\ref{app:feature_selection}.

\noindent\textbf{Literal-transcription diagnostic.} After inspecting the translation gap, we add literal Chinese transcriptions to all twenty image pairs, retaining the images and questions and adding no English translations or answer labels. Scores rise from 100.00 to 177.50 for 7B and from 55.00 to 155.00 for 13B. Paired gains are 77.50 points ($[37.50,117.50]$) and 100.00 ($[67.50,130.00]$); the assisted 13B-minus-7B interval is $[-67.50,22.50]$. Direct textual access recovers many decisions, without separating visual recognition from use of visual features or establishing a training-data cause. These post-hoc assisted scores remain separate from benchmark performance.

\FloatBarrier
\section{Training Progress at Fixed Model Size}
\label{app:training_progress}

\subsection{Matched low-data loss trajectories}

\begin{figure*}[!ht]
\centering
\includegraphics[width=\textwidth]{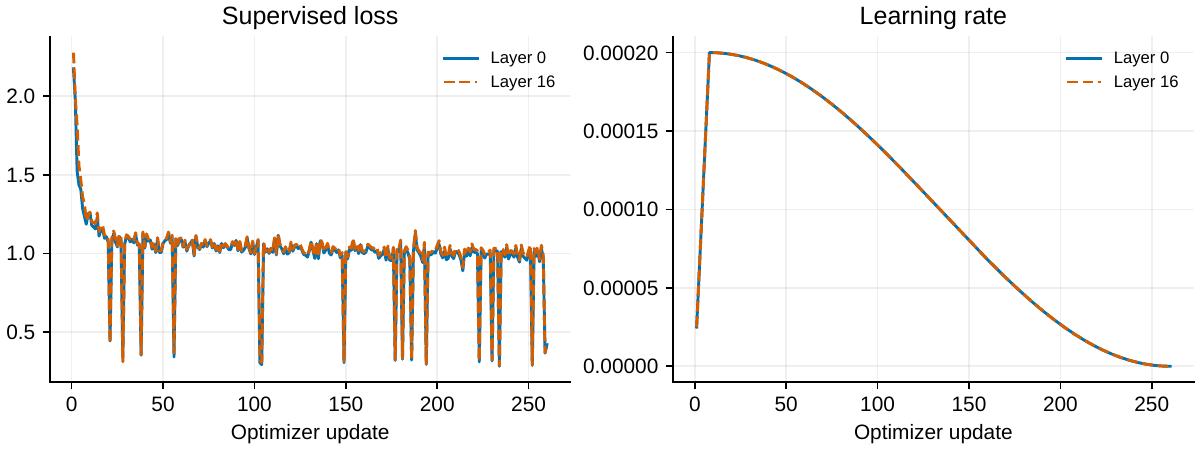}
\caption{Matched Inverse-LLaVA depth study on 5\% of the instruction data.
Raw token-weighted training loss and recorded learning rate are shown at
every optimizer update, without smoothing. Both models exclude LoRA from
layers 0 and 16 to match adapter capacity. The horizontal axis describes
optimization progress within these runs.}
\label{fig:depth_loss}
\end{figure*}

Figure~\ref{fig:depth_loss} overlays all 260 recorded updates from the matched
layer-0/layer-16 study on the same 5\% instruction subset. Both runs share
selected row IDs, local batch partition, target counts, learning-rate schedule,
and LoRA exclusion mask. The final 20-update mean training losses are 0.8962
and 0.9129, respectively. Their final OCRBench scores are both 2.20\%, while
MME cognition scores are 289.29 and 287.86. The loss separation describes
optimization on these training batches; broader task rankings require
checkpoint evaluations on non-floor endpoints.

\FloatBarrier

\subsection{Full-run checkpoint evaluations}

\begin{figure*}[!ht]
\centering
\includegraphics[width=\textwidth]{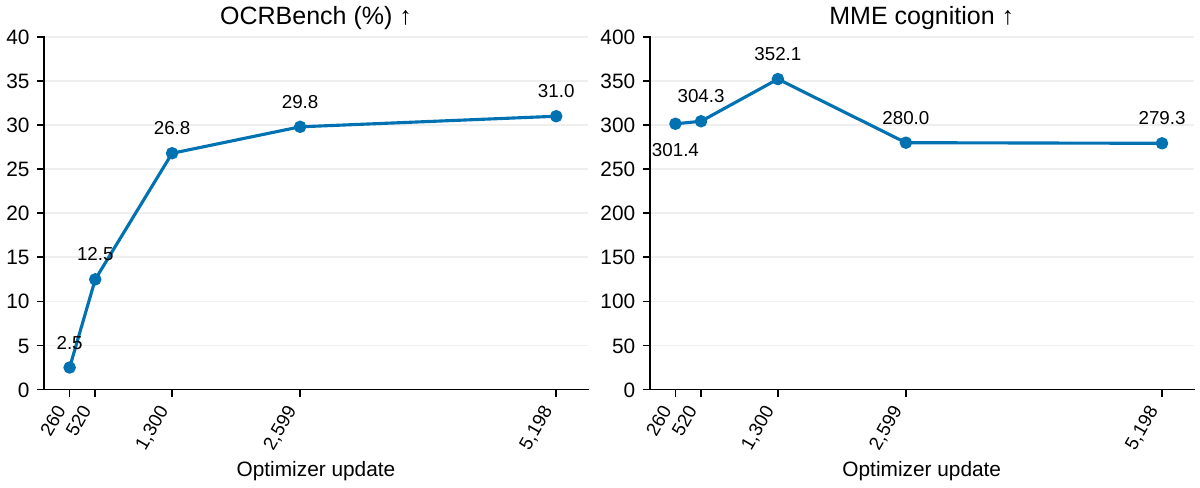}
\caption{Complete OCRBench and MME cognition evaluations at five retained optimizer updates of one seed-42 training run. Both panels use identical update coordinates. With global batch 128, update 5,198 corresponds to 665,344 sampled rows: 665,298 training rows plus 46 padding repetitions. The milestones cover approximately 5\%, 10\%, 25\%, 50\%, and 100\% of the run. OCR improves with training, while cognition follows a different trajectory. Every point has a verified checkpoint and score record.}
\label{fig:training_progress}
\end{figure*}

Figure~\ref{fig:training_progress} follows five declared checkpoints from one 7B instruction-tuning run. OCRBench rises from 2.5\% at update 260 to 12.5\%, 26.8\%, 29.8\%, and 31.0\% at updates 520, 1300, 2599, and 5198. The shared schedule couples sample exposure, optimization progress, and learning rate. Dedicated data-size and backbone comparisons are required for scaling claims.

MME cognition peaks at 352.14 at update 1300 and reaches 279.29 at completion. Translation falls from 122.5 to 57.5; the final checkpoint answers 39 of 40 translation questions ``No.'' Numerical calculation ranges from 42.5 to 57.5. These response patterns motivate supervision and retention tests. The final checkpoint remains primary; the intermediate peak is descriptive and was not used for selection.

\FloatBarrier

\section{Matched Instruction Replay}
\label{app:instruction_replay}

The parent checkpoint and its paired-1\% descendant each receive the same source-stratified 1\% instruction subset: 6,653 unique rows, 6,656 sampled rows after padding, and 845,999 surviving supervised targets. Both jointly train LoRA and fusion for 52 updates with fresh optimizers, seed 42, data seed 17, global batch 128, peak learning rate $2\times10^{-5}$, and the same one-epoch cosine schedule. These are continuations of instruction-trained checkpoints; the experiment does not test paired pretraining from a fresh initialization.

Table~\ref{tab:instruction_replay} retains all four checkpoints; the following analysis reports the planned paired contrasts. Evaluation covers complete OCRBench (1,000 questions), TextVQA validation (5,000), ScienceQA-IMG (2,017), and the stated MME cognition and celebrity panels (260 and 340). All four checkpoints use H100 with BF16, batch-one greedy decoding, unmerged adapters, and explicit flash-only SDPA. Official prompts, generation limits, and scorers are unchanged. These are exploratory evaluations of previously consulted benchmarks. All immediate parents are re-evaluated under this policy, so each intervention is compared with its own matched control; Table~\ref{tab:recovery_panel} uses the primary Thor policy. Every response in the earlier fixed 256-question screens matches its full-run counterpart exactly. Full-split scores take priority over subset estimates.

After replay, ScienceQA accuracy rises by 13.98 points (95\% paired-image interval $[12.20,15.78]$) and invalid answers fall from 457 to zero. Of those invalid answers, 448 are standalone lowercase choices. The case-only diagnostic scores 67.77\%, compared with 68.17\% after replay; much of the recovery concerns response format. The instruction-only control declines from 70.00\% to 68.77\% ($-1.24$ points, $[-2.05,-0.43]$).

TextVQA improves by 2.25 points ($[1.47,3.04]$) from the paired-only parent. OCRBench changes by $+1.00$ ($[-0.49,2.44]$). Comparing the two replayed models gives $-0.30$ OCRBench points ($[-1.21,0.60]$), $-0.26$ TextVQA points ($[-0.76,0.25]$), and $-0.59$ ScienceQA points ($[-1.28,0.10]$); each interval includes zero. MME cognition increases by 32.50 points ($[8.21,60.71]$) relative to the replayed control, while celebrity changes by $+4.12$ ($[-1.18,9.71]$). Celebrity falls by 15.29 points from the paired-only parent ($[-25.00,-6.18]$). The intervention therefore restores instruction-following behavior while reducing the earlier celebrity gain. All intervals are conditional on fitted checkpoints and unadjusted for multiplicity. All four models use inverse fusion, so this experiment tests supervision sequence rather than architecture.

An exploratory TextVQA audit finds similar final replay gains when a supported reference appears in the supplied OCR hints ($+2.18$ points; 2,355 questions) or is absent ($+2.31$; 2,645 questions). A supported reference is given by at least two annotators. Mean normalized answer length increases from 1.45 to 1.50 words, while reference-containing longer responses decrease from 319 to 236. These descriptive changes motivate examining answer selection beyond length alone; they do not isolate visual recognition or change official scores.

\begin{table*}[!ht]
\centering
\small
\caption{Matched instruction replay. Complete OCRBench, TextVQA, ScienceQA-IMG, and stated MME panels. Accuracy metrics are percentages; MME uses score points. All four checkpoints use the same H100 execution policy and official scorers.}
\label{tab:instruction_replay}
\setlength{\tabcolsep}{4pt}
\begin{tabular}{@{}lrrrrr@{}}
\toprule
Checkpoint & OCRBench & TextVQA & SQA & MME cog. & Celebrity \\
\midrule
Inverse-LLaVA & 30.50 & 56.95 & 70.00 & 272.14 & 127.06 \\
Paired 1\% & 29.50 & 54.32 & 54.19 & 325.36 & 152.94 \\
\midrule
Parent + instruction 1\% & 30.80 & 56.83 & 68.77 & 271.79 & 133.53 \\
Paired 1\% + instruction 1\% & 30.50 & 56.58 & 68.17 & 304.29 & 137.65 \\
\bottomrule
\end{tabular}
\par\smallskip\begin{minipage}{0.98\linewidth}\footnotesize Upper block: parent checkpoints. Lower block: both parents after the same instruction replay.\end{minipage}
\end{table*}

\noindent\textbf{Transfer to MMBench.} Table~\ref{tab:mmbench_transfer} evaluates six retained checkpoints on complete EN dev. Our extractor matches the pinned upstream code on all 25,974 responses. The paired 1\%/5\% branches produce 1,509/3,342 standalone lowercase labels. Case-only diagnostics reach 55.76\%/26.46\%, respectively, both below the parent's 62.63\%. Declared scores remain unchanged.

Instruction replay raises the paired-1\% score from 29.90\% to 61.60\%, while the equally replayed parent also scores 61.60\% (paired difference $0.00$ points, 95\% interval $[-0.95,0.95]$). The paired-then-replayed model remains 1.03 points below the unchanged parent ($[-1.98,-0.09]$). Thus, replay recovers most deterioration without establishing an additional overall benefit from paired supervision. All eight prespecified contrasts and all twenty categories are retained; intervals are conditional on these checkpoints and unadjusted for multiplicity.

\begin{table*}[!ht]
\centering\small
\caption{MMBench EN supervision transfer on 1,164 circular groups (4,329 rotations), with identical released LLaVA prompts and deterministic extraction. Unparsed counts refer to rotations. The case-only column is a posthoc diagnostic: only unparsed single-letter valid options are uppercased, without consulting references. It does not replace the declared benchmark score.}
\label{tab:mmbench_transfer}
\setlength{\tabcolsep}{5pt}
\begin{tabular}{@{}lrrr@{}}
\toprule
Checkpoint & \shortstack{Circular\\accuracy (\%)} & Unparsed & \shortstack{Case-only\\diagnostic (\%)} \\
\midrule
\rowcolor{methodshade}
Inverse-LLaVA & 62.63 & 0 & 62.63 \\
Paired 1\% & 29.90 & 1,590 & 55.76 \\
Paired 5\% & 1.46 & 4,028 & 26.46 \\
Instruction (token control) & 62.37 & 0 & 62.37 \\
Parent + instruction 1\% & 61.60 & 0 & 61.60 \\
Paired 1\% + instruction 1\% & 61.60 & 1 & 61.60 \\
\bottomrule
\end{tabular}
\par\smallskip\begin{minipage}{0.98\linewidth}\footnotesize Gray marks the unchanged Inverse-LLaVA parent; comparisons use the stated controls.\end{minipage}
\end{table*}

\begin{figure*}[!ht]
\centering
\includegraphics[width=\textwidth]{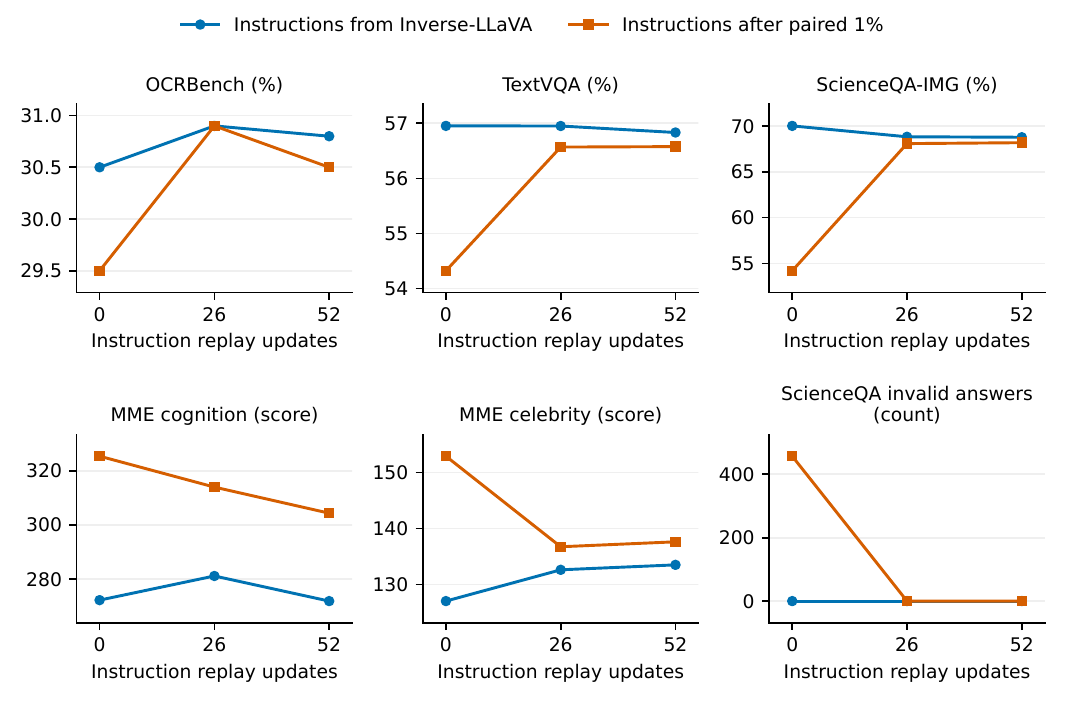}
\caption{Saved instruction-replay trajectories on the five complete evaluation panels, with official ScienceQA invalid-answer counts. Both branches receive the same ordered instructions under the same 52-update schedule. Points are observations at updates 0, 26, and 52; connecting lines guide the eye. Vertical scales differ across panels. This is a single-seed trajectory diagnostic, without checkpoint selection or a fitted data-scaling law.}
\label{fig:replay_trajectory}
\end{figure*}

\noindent\textbf{Midpoint evaluation.} Figure~\ref{fig:replay_trajectory} evaluates both update-26 checkpoints, each trained on 3,328 sampled instruction examples and 456,027 supervised targets. All 457 invalid ScienceQA outputs have disappeared by this midpoint. For the paired branch, TextVQA rises from 54.32\% to 56.57\% ($+2.24$ points, 95\% paired interval $[1.46,3.03]$); updates 26--52 add $0.01$ points ($[-0.23,0.25]$). Celebrity decreases by 16.18 score points by update 26 ($[-25.59,-7.06]$), followed by a change of $+0.88$ ($[-2.65,4.41]$). Both midpoint OCRBench scores are 30.9\% (difference $0.00$ points, $[-0.91,0.92]$). Most measured change occurs by update 26. The final comparison still uses update 52; all 25 paired contrasts are reported with conditional, unadjusted intervals.

\begin{figure*}[!ht]
\centering
\includegraphics[width=\textwidth]{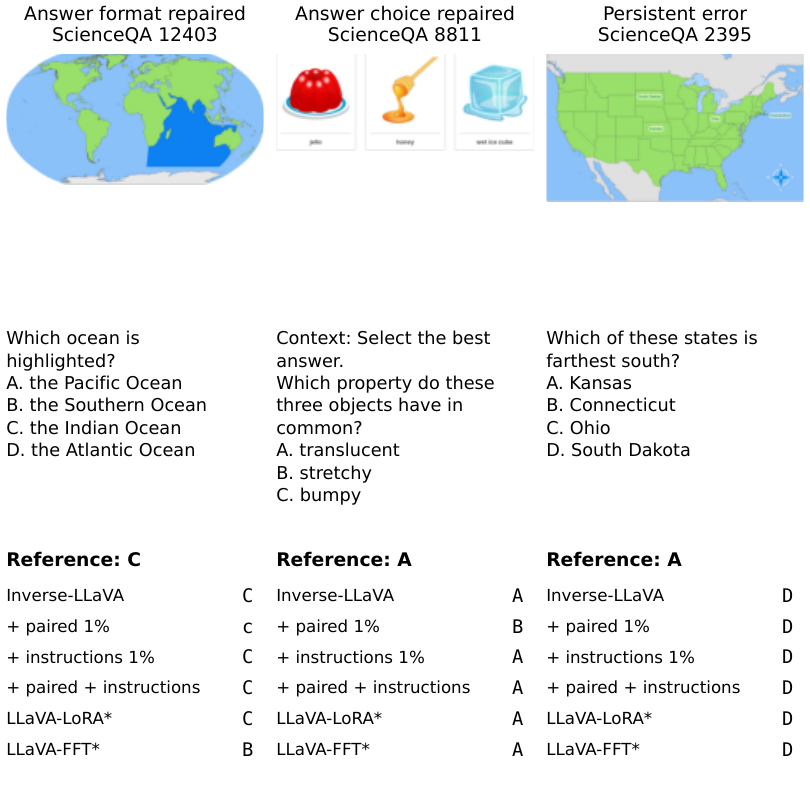}
\caption{ScienceQA~\citep{lu2022scienceqa} format repair, corrected choice, and persistent error, selected by first hash rank within each outcome group of the fixed screen. The complete ten-case record retains regressions and consistent successes. Answers are unedited, including letter case. All additions start from Inverse-LLaVA; paired and instruction doses are each 1\%. The four treatment/control outputs use the matched H100 policy. Asterisks identify official LLaVA-LoRA and FFT responses retained from Thor as separate qualitative context. Questions omit only the chat wrapper and fixed option-letter instruction. These outcome-selected illustrations document response changes without estimating their frequency or assigning an internal reasoning mechanism.}
\label{fig:replay_cases}
\end{figure*}

\FloatBarrier
\section{Training and Evaluation Details}
\label{app:reproduction_protocols}

\noindent\textbf{Sample ordering.} An epoch-seeded generator groups examples by modality and length in 128-row megabatches, balancing their lengths across four consecutive 32-row batches. Incomplete modality remainders are combined and length-balanced before distributed tail padding. This generator is isolated from the model's random state. Controlled subset comparisons use the same ordering and local batch partition, with per-update exposure audits.

\noindent\textbf{Training data and execution.} Grouped by image family, the normalized instruction mixture contains 364,100 COCO rows, 86,417 Visual Genome rows, 72,140 GQA rows, 80,000 OCR-VQA rows, 21,953 TextCaps caption rows stored under the TextVQA image directory, and 40,688 text-only rows. Image-family counts do not identify each supervision task: the COCO family, for example, serves several tasks. All 349,034 unique referenced images pass decoding checks. Deterministic row-qualified identifiers distinguish repeated source IDs. Following the Vicuna-v1 masking policy, delimiter-ambiguous rows retain their input context with all targets masked; an entirely unsupervised expanded microbatch is rejected. The one-epoch training run has 5,198 optimizer updates at global batch 128; the sampler pads its final distributed batch. Seed 42 fixes initialization and sampling. AdamW uses $\beta=(0.9,0.999)$, $\epsilon=10^{-8}$, and gradient clipping at norm 1.0. Trainable master parameters and optimizer moments are FP32, with BF16 model computation. Checkpoints retain optimizer, scheduler, sampler progress, and per-rank random states to restore the recorded training state. The disclosed numerical policy does not guarantee bitwise replay.

\noindent\textbf{Initialization.} Mapper entries are sampled from $\mathcal{N}(0,1/d_v)$ and fusion-output entries from $\mathcal{N}(0,10^{-8})$, where the second argument denotes variance. Visual RMSNorm gains and fusion scales start at one; visual RMSNorm uses $\epsilon=10^{-6}$. LoRA initializes its input factor with Kaiming-uniform weights and its output factor to zero. Fresh tensors are initialized in FP32 and cast to the model dtype before training.

\noindent\textbf{Training-time accounting.} Logged intervals for updates 1--1,300 and 1,301--5,198 total 17,606.65 and 53,475.93 seconds, giving 19.75 hours. Including two discarded attempts gives 21.21 logged hours. Setup, unlogged failures, session gaps, and terminal checkpoint writes are excluded from both totals; the logs retain accepted and discarded updates separately. The 665,344 sampled rows include 46 padding repeats.

\noindent\textbf{Parameter accounting.} Total/trainable counts are 7,389,496,323/347,573,251 for Inverse-LLaVA, 7,382,718,464/340,795,392 for official LoRA, and 7,062,902,784/6,759,395,328 for original FFT under frozen-CLIP training. Inverse-LLaVA has 6,777,859 more trainable parameters than LoRA. The evaluated HF-converted FFT checkpoint has 7,063,427,072 total parameters: 64 additional rows in each untied embedding and output head add 524,288 parameters. Training-policy counts are independent of inference-time gradient flags.

\noindent\textbf{Supervision audit.} Auditing all 665,298 normalized rows at context length 2,048 and 576 image patches finds no row-contract errors. All 2,638,095 OCR-VQA targets and 413,765 TextCaps caption targets survive collation and image expansion. Seven text-only rows have no supervised targets under the declared masking policy. Thus, target truncation does not explain the OCR deficit for these sources under this recipe. These unique-row counts do not measure sampled epoch weighting or establish the effectiveness of additional OCR exposure.

\noindent\textbf{Image overlap.} Exact decoded-RGB comparisons find no shared images between the eight local evaluation sets and either the complete instruction or paired-data adaptation pool. These checks bind the evaluated image manifests to the training inventories. They do not establish absence of near-duplicates, text-only leakage, or overlap with CLIP/Vicuna pretraining data.

\noindent\textbf{Component response patterns.} Addition answers every numerical-calculation and code-reasoning question ``No,'' yielding no fully correct pairs. Concatenation and gating also answer every numerical question ``No.'' Gating answers 38 of 40 code questions ``No,'' with two fully correct pairs. These category-specific response patterns qualify the aggregate component scores.

\noindent\textbf{Matched depth training.} The layer-0 and layer-16 conditions retain all 32 decoder layers and exclude both layers 0 and 16 from LoRA adaptation. Each has 337,579,011 trainable parameters. They use identical ordered selections of 33,265 instruction rows, padded to 33,280 sampled rows over 260 updates, and the same optimizer and batch configuration. The audited per-update sample, expanded-token, and supervised-target counts match; each run consumes 4,209,792 supervised targets. This comparison is separate from the component screen, whose LoRA placement differs.

Both models score 2.20\% on full OCRBench; the layer-16-minus-layer-0 difference is 0.00 percentage points (95\% paired interval $[-0.81,0.82]$). MME cognition scores are 289.29 and 287.86, giving a difference of $-1.43$ points ($[-48.22,46.79]$). Both answer every numerical-calculation question ``No.'' The two tested depths have close point scores under this limited-data recipe, but the low OCR scores and wide cognition interval prevent a conclusion of full-data insensitivity.

\noindent\textbf{Continuation exposure.} Each recovery branch starts independently from the final Inverse-LLaVA checkpoint and jointly trains LoRA and fusion parameters with a fresh optimizer. The six main conditions use global batch 128 and peak learning rate $2\times10^{-5}$. Paired 1\% and 5\% are nominal doses of 5,580 and 27,900 unique rows from the verified 558,128-pair pool. OCR replay and the instruction/non-OCR controls draw from the instruction mixture. Table~\ref{tab:recovery_exposure} reports actual exposure, including sampler padding. Approximate token matching does not match optimizer updates, schedule length, or source composition; the row- and token-matched controls therefore answer complementary questions. The separate 128-row optimization diagnostic repeats its examples for three epochs and is excluded from the primary generalization panel. Training completion and falling training loss alone do not establish held-out recovery.

\begin{table*}[!ht]
\centering
\small
\caption{Completed continuation exposures from the same parent checkpoint. Targets count surviving supervised tokens over sampled rows. The instruction-token control approximates paired-1\% targets; the non-OCR token control approximates OCR-replay targets.}
\label{tab:recovery_exposure}
\begin{tabular}{lrrrr}
\toprule
Condition & Unique rows & Sampled rows & Updates & Targets \\
\midrule
Paired 1\% & 5,580 & 5,632 & 44 & 84,740 \\
Paired 5\% & 27,900 & 27,904 & 218 & 422,883 \\
OCR replay & 5,580 & 5,632 & 44 & 168,882 \\
Instruction (tokens) & 692 & 704 & 6 & 86,469 \\
Non-OCR (rows) & 5,580 & 5,632 & 44 & 616,012 \\
Non-OCR (tokens) & 1,498 & 1,536 & 12 & 173,520 \\
\midrule
Optimization diagnostic & 128 & 384 & 48 & 12,000 \\
\bottomrule
\end{tabular}
\end{table*}

\noindent\textbf{Schedule convention.} The training schedule rounds warmup length to $w=\operatorname{round}(0.03T)$ for $T$ optimizer updates. The multiplier is $\max(s,1)/w$ for $0\leq s<w$, followed by cosine decay, where $s$ counts completed updates. The full run uses 156 warmup updates and an initial learning rate of approximately $1.282\times10^{-6}$. Component screens share this implementation.

\begin{table*}[!ht]
\centering
\caption{Benchmark evaluation protocols. Limits are maximum generated tokens; generation can stop earlier at EOS. Every row uses the complete stated split.}
\label{tab:local_protocols}
\small
\begin{tabular}{>{\raggedright\arraybackslash}p{2.3cm}r>{\raggedright\arraybackslash}p{6.4cm}r}
\toprule
Dataset and split & Questions & Input and scoring rule & Limit \\
\midrule
OCRBench test & 1,000 & Released question; category-specific OCRBench normalization and matching. & 128 \\
MME cognition & 260 & Released LLaVA question with short-answer suffix; category sum of accuracy and paired-image accuracy. & 128 \\
MME perception & 2,114 & Same MME protocol, preserving both questions per image and all perception categories. & 128 \\
TextVQA validation & 5,000 & Version 0.5.1 questions with Rosetta v0.2 OCR tokens; official answer normalization and ten-annotator consensus. & 1,024 \\
ScienceQA-IMG test & 2,017 & Released LLaVA CQM-A inputs, restricted to image-bearing questions; official option extraction and accuracy. & 1,024 \\
AI2D test & 3,088 & Question and choices with letter-only instruction; pinned response filter and exact-match accuracy. & 16 \\
MMStar validation & 1,500 & Released choices and letter-only instruction; pinned option extraction and macro-L2 aggregation. & 128 \\
GQA test-dev balanced & 12,578 & Released LLaVA short-answer prompt; exact normalized answer accuracy on public labels. & 128 \\
VizWiz test & 8,000 & Released LLaVA questions and explicit unanswerable instruction; April 2026 public test references and official consensus scoring. & 128 \\
VQAv2 test-dev & 107,394 & Released LLaVA short-answer prompts and M4C normalization; official hidden-label test-dev scoring. & 128 \\
MM-Vet v1 & 218 & Released LLaVA prompts; official hosted GPT-4.1 (2025-04-14), one grading run per model; raw item grades retained. & 1,024 \\
MMBench EN/CN dev & 4,329 each & Released language-specific LLaVA prompts; 1,164 circular question groups per language; deterministic extraction with no judge fallback. & 1,024 \\
\bottomrule
\end{tabular}
\end{table*}

\noindent\textbf{Numerical and protocol controls.} Primary predictions use batch-one greedy decoding, one beam, KV caching, BF16, and SDPA. Inverse and LoRA use unmerged adapters unless specified; FFT uses Transformers generation. Table~\ref{tab:canonical_panel}, including OCRBench, uses NVIDIA Thor; additional original-benchmark evaluations use one H100 SXM with flash-only SDPA. Each benchmark uses a common device policy for all three models. Snapshot inventories verify identical FFT weights across hosts; predictions retain runtime settings. ScienceQA IDs and prompts match released CQM-A inputs row-for-row. Scorer checks cover extraction, aggregation, and complete coverage. Public-label scores retain their split identity and cannot replace hidden-label submissions.

\begin{samepage}
\noindent\textbf{VQAv2 submission.} We follow LLaVA's test-dev procedure: generate all 107,394 test-dev answers, apply its released normalization, and construct the 447,793-row submission envelope with empty answers for the remaining IDs. The complete upload is checked against the pinned LLaVA converter and submitted to EvalAI challenge 830, test-dev phase 1793. The reported score covers test-dev only; this envelope does not establish test-standard performance. Server results provide aggregate accuracy by answer type, without the itemwise scores needed for paired bootstrap intervals.\par
\end{samepage}

\begin{samepage}
\noindent\textbf{Uncertainty.} Reported 95\% comparison intervals use 10,000 paired bootstrap resamples with seed 2026. OCRBench, TextVQA, ScienceQA, AI2D, GQA, VizWiz, and MM-Vet resample clusters of questions sharing identical decoded RGB image content, retaining question-weighted scores when cluster sizes differ. MM-Vet contains 199 such groups across 200 image files; its intervals condition on the retained judge outputs. MMBench resamples complete circular question groups. MME retains its official paired-question and category aggregation; MMStar retains fixed category strata and macro aggregation. These intervals quantify benchmark-sample uncertainty conditional on the fitted checkpoints; they do not quantify training-seed variability. Qualitative outcome-balanced selections and representation interventions remain separate from benchmark accuracy estimates.
\par
\end{samepage}
\FloatBarrier
\section{Text-Only Evaluation Protocol}
\label{app:language_retention}

\begin{table}[!ht]
\centering
\small
\caption{Text-only normalized accuracy (\%) on complete ARC-Challenge and HellaSwag splits, using 25-shot and zero-shot evaluation, respectively.}
\label{tab:language_retention}
\begin{tabular}{lrr}
\toprule
Model & ARC-Challenge & HellaSwag \\
\midrule
\rowcolor{methodshade}
Inverse-LLaVA & \underline{53.41} & \underline{73.18} \\
LLaVA-LoRA & 53.24 & 71.44 \\
LLaVA-FFT & 52.90 & \textbf{74.46} \\
Vicuna-7B & \textbf{53.84} & 71.58 \\
\bottomrule
\end{tabular}
\par\smallskip\begin{minipage}{0.98\linewidth}\footnotesize Bold/underlined: highest/second-highest score per metric. Gray identifies Inverse-LLaVA.\end{minipage}
\end{table}

We use \texttt{lm-eval} 0.4.12 at source revision \texttt{6d642546f468}, with seed 42, raw causal prompts, explicit BOS, and a 2,048-token context. ARC-Challenge~\citep{clark2018arc} includes all 1,172 questions with 25 demonstrations; HellaSwag~\citep{zellers2019hellaswag} includes all 10,042 with none. Both report length-normalized choice accuracy. Evaluation uses BF16 and batch size four on Thor, without images or a chat template; adapters remain unmerged. Token-boundary, padding, and left-truncation checks agree with the pinned HFLM interface. Predictions retain document, prompt, and target hashes.

All six comparisons use 10,000 paired-item resamples, seed 2026, after matching those hashes. Inverse minus LoRA, Vicuna, and FFT on ARC-Challenge is respectively $+0.17$ ($[-1.71,2.13]$), $-0.43$ ($[-2.30,1.45]$), and $+0.51$ ($[-1.37,2.39]$) percentage points. These intervals do not establish equivalence. HellaSwag comparisons are reported in the main text. The intervals are descriptive and unadjusted for multiplicity; they condition on these fitted checkpoints and exclude training-seed variation. This two-task panel does not establish retention across all language capabilities.
\begin{samepage}
\section{Supervision Effects by Category}
\label{app:recovery_categories}

These figures add paired intervals and MME perception categories to
Table~\ref{tab:recovery_panel}.\par
\end{samepage}

\begin{figure}[H]
\centering
\includegraphics[width=\textwidth]{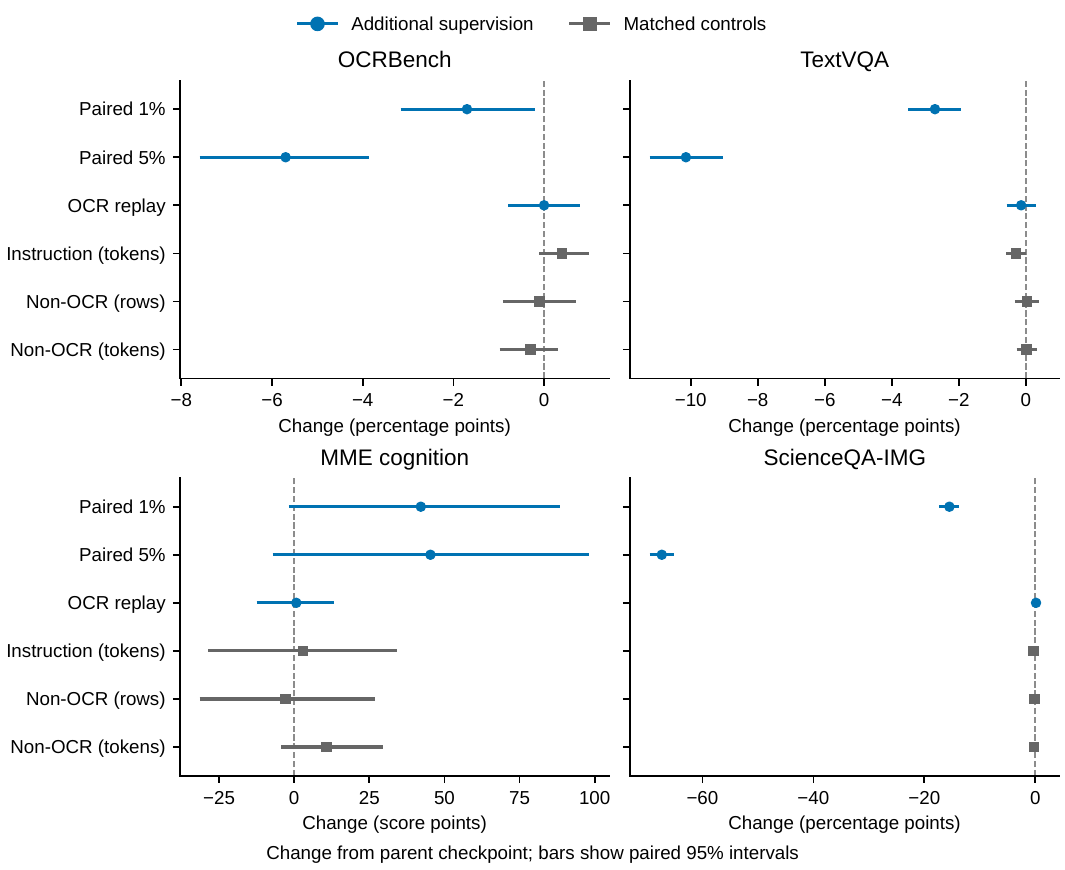}
\caption{Continuation effects relative to the Inverse-LLaVA parent. Markers show
paired differences and lines show 95\% bootstrap intervals, conditional on the
fitted checkpoints and unadjusted for multiplicity. All conditions are shown.
Accuracy panels use percentage points; MME cognition uses score points.
ScienceQA retains case-sensitive scoring and therefore includes response-format failures.}
\label{fig:recovery_effects}
\end{figure}

\begin{figure*}[!p]
\centering
\includegraphics[width=\textwidth]{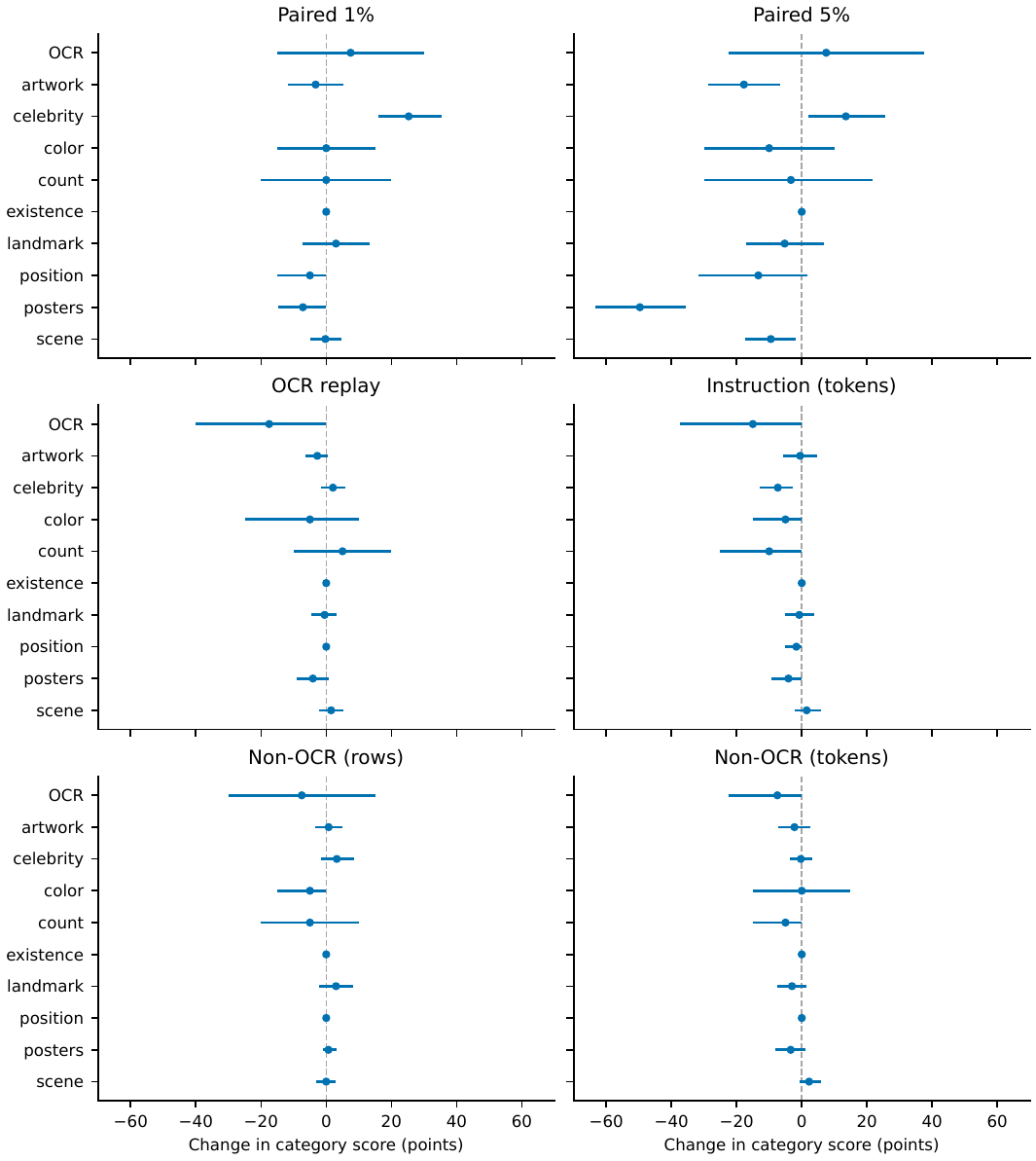}
\caption{MME perception category changes from the same Inverse-LLaVA parent checkpoint.
All ten categories and six continuations share a horizontal scale. Dots show
native score differences; lines give 95\% intervals from 10,000 paired
image-group bootstrap resamples (seed 2026). Each category has a maximum of 200.
Intervals condition on the fitted checkpoints, are unadjusted for multiple
comparisons, and exclude training-seed variation. Zero-width intervals indicate
identical paired correctness in this sample, without establishing population equivalence.
Paired-data continuations improve celebrity recognition under the tested
recipes; the 5\% dose also lowers artwork, poster, and scene scores.
Aggregate and other-task effects are reported separately.}
\label{fig:recovery_perception_categories}
\end{figure*}

\begin{figure*}[!p]
\centering
\includegraphics[width=\textwidth]{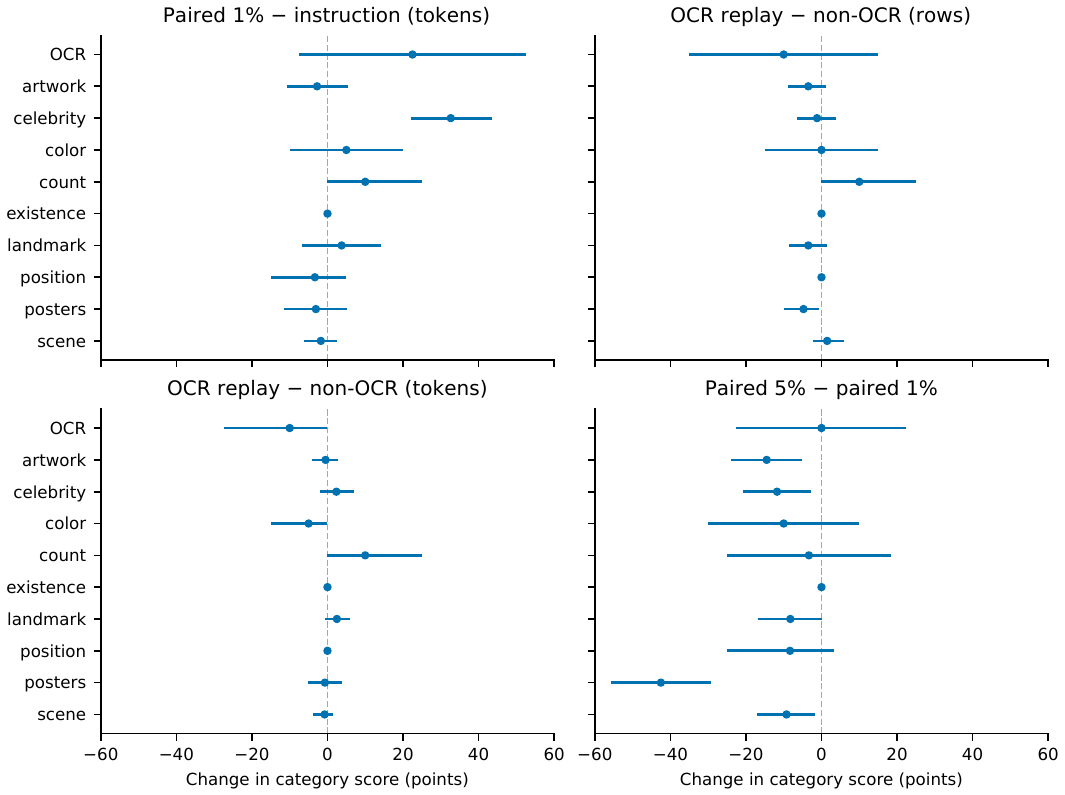}
\caption{All four declared direct continuation contrasts, with the direction
specified in each panel title. All ten categories are retained, using the
interval procedure and interpretation in Figure~\ref{fig:recovery_perception_categories}.
Row- and token-matched controls assess
complementary exposure differences; approximate token matching does not match
update count, schedule length, or source composition. The paired-1\% celebrity
gain also appears relative to the instruction-token control. These comparisons
do not isolate architecture from supervision or establish broad OCR recovery.}
\label{fig:recovery_perception_controls}
\end{figure*}
\end{appendices}

\clearpage
\bibliography{references}

\end{document}